\documentclass[sigconf]{acmart}
\usepackage{multirow}
\usepackage{graphicx}
\usepackage{multirow}
\usepackage{booktabs}
\usepackage{enumitem}
\usepackage{xcolor}
\usepackage{pifont}
\usepackage{lineno}
\usepackage{xspace}

\usepackage{amsmath,amsfonts,bm}

\def\eqref#1{equation~\ref{#1}}

\def\1{\bm{1}}

\def\rmX{{\mathbf{X}}}

\DeclareMathAlphabet{\mathsfit}{\encodingdefault}{\sfdefault}{m}{sl}
\SetMathAlphabet{\mathsfit}{bold}{\encodingdefault}{\sfdefault}{bx}{n}

\def\gA{{\mathcal{A}}}

\def\gE{{\mathcal{E}}}

\def\gG{{\mathcal{G}}}
\def\gH{{\mathcal{H}}}

\def\gR{{\mathcal{R}}}

\def\gV{{\mathcal{V}}}

\def\gX{{\mathcal{X}}}
\def\gY{{\mathcal{Y}}}

\definecolor{darkgreen}{RGB}{0, 100, 0}
\definecolor{lightgreen}{RGB}{144, 238, 144}

\newtheoremstyle{noIndentStyle} 
  {3pt} 
  {3pt} 
  {} 
  {} 
  {\bfseries} 
  {.} 
  { } 
  {\thmname{#1}\thmnumber{ #2}\thmnote{ (#3)}} 

\theoremstyle{noIndentStyle}

\newcommand{\modelname}{\textsc{Opbench}\xspace}
\newcommand{\het}{\ensuremath{\gG_\text{het}}\xspace}
\newcommand{\hp}{\ensuremath{\gH}\xspace}

\newcommand\blfootnote[1]{%
  \begingroup
  \renewcommand\thefootnote{}\footnote{#1}%
  \addtocounter{footnote}{-1}%
  \endgroup
}

\AtBeginDocument{%
  }

\setcopyright{acmcopyright}
\usepackage{hyperref}
\copyrightyear{2026} 
\acmYear{2026} 
\setcopyright{rightsretained} 
\acmConference[KDD '26]{Proceedings of the 32th ACM SIGKDD Conference on Knowledge Discovery and Data Mining}{August 9--13, 2026}{Jeju, Korea}
\acmBooktitle{Proceedings of the 32th ACM SIGKDD Conference on Knowledge Discovery and Data Mining (KDD '26), August 9--13, 2026, Jeju, Korea}

\begin{document}

\title{\modelname: A Graph Benchmark to Combat the Opioid Crisis}
\definecolor{customizedgrey}{gray}{0.25}

\author{
\textbf{Tianyi Ma\textsuperscript{1}},
\textbf{Yiyang Li\textsuperscript{1}},
\textbf{Yiyue Qian\textsuperscript{3}\textsuperscript{$\dagger$}},
\textbf{Zheyuan Zhang\textsuperscript{1}},
\textbf{Zehong Wang\textsuperscript{1}},\\
\textbf{Chuxu Zhang\textsuperscript{2}}, and
\textbf{Yanfang Ye\textsuperscript{1}\textsuperscript{$\ddagger$}}\\ 
{\textit{\textsuperscript{1}University of Notre Dame, \textsuperscript{2}University of Connecticut, \textsuperscript{3}Amazon}}\\
{\{tma2, yli62\}@nd.edu, yyqian5@gmail.com, \{zzhang42, zwang43\}@nd.edu,\\ 
chuxu.zhang@uconn.edu, yye7@nd.edu}\\   
}

\renewcommand{\shortauthors}{Tianyi Ma et al.}
\begin{abstract}
The opioid epidemic continues to ravage communities worldwide, straining healthcare systems, disrupting families, and demanding urgent computational solutions.
To combat this lethal opioid crisis, graph learning methods have emerged as a promising paradigm for modeling complex drug-related phenomena. 
However, a significant gap remains: there is no comprehensive benchmark for systematically evaluating these methods across real-world opioid crisis scenarios.
This gap persists due to three key challenges: 
(i) the multidimensional nature of the crisis spanning the supply and demand sides of the drug ecosystem, 
(ii) the structural complexity of drug-related data involving heterogeneous and higher-order interactions, and 
(iii) the scarcity of publicly available annotated datasets due to privacy constraints and labor-intensive expert labeling.
To bridge this gap, we introduce \modelname, the first comprehensive opioid benchmark comprising five datasets across three critical application domains: opioid overdose detection from healthcare claims, illicit drug trafficking detection from digital platforms, and drug misuse prediction from dietary patterns.
Specifically, 
\modelname incorporates diverse graph structures, including heterogeneous graphs and hypergraphs, to preserve the rich and complex relational information among drug-related data.
To address data scarcity, we collaborate with domain experts and authoritative institutions to curate and annotate datasets while adhering to privacy and ethical guidelines.
Furthermore, we establish a unified evaluation framework with standardized protocols, predefined data splits, and reproducible baselines to facilitate fair and systematic comparison among graph learning methods.
Through extensive experiments, we analyze the strengths and limitations of existing graph learning methods, thereby providing actionable insights for future research in combating the opioid crisis.
Our source code and datasets are available at \href{https://github.com/Tianyi-Billy-Ma/OPBench}{https://github.com/Tianyi-Billy-Ma/OPBench}.
\end{abstract}

\maketitle
\blfootnote{$\dagger$ The work is not related to the position at the corresponding institution.}
\blfootnote{$\ddagger$ Corresponding Author.}

\section{Introduction}

The global opioid crisis has become one of the most devastating public health emergencies, exacting an immense toll on human lives, healthcare systems, and socioeconomic stability~\cite{who2023urgent}.
The epidemic imposes a substantial economic burden, costing an estimated \$1.5 trillion per year in healthcare spending, lost productivity, and criminal justice expenses~\cite{florence2021economic, maclean2020economic}.
Beyond economic costs, the crisis profoundly disrupts families and communities, with nearly 105,000 individuals dying from opioid overdoses in 2023, representing a staggering tenfold increase from 1999~\cite{cdc_understanding_opioid_overdose_epidemic, brundage2019ripple}, which calls for imminent actions to address this issue.
To combat the devastating and lethal opioid crisis, graph learning methods~\cite{fan2017social,kamdar2019knowledge, fan2018automatic, wang2026molecular, zhang2025ngqa, ma2023hypergraph} have emerged as a compelling paradigm, owing to their capacity in learning latent structural patterns and rich relational information inherent in complex real-world systems~\cite{wang2025generative}.
For example, DDHGNN~\cite{wen2022disentangled} proposes a dynamic heterogeneous graph neural network that models interactions among patients, drugs, physicians, and pharmacies over time to identify patients at risk of overprescription and potential overdose, demonstrating the strong capabilities of graph learning methods for combating the opioid crisis.

Although graph learning methods have shown promising progress in combating the opioid crisis, no comprehensive benchmark exists to systematically evaluate and compare these methods across real-world scenarios.
This gap persists due to three key challenges.
(i) The opioid crisis is inherently \textbf{multidimensional}, spanning both the \textbf{supply and demand sides of the drug ecosystem}, requiring benchmarks that cover diverse application scenarios. 
On the supply side, opioid over-prescription in healthcare systems places patients at heightened risk of addiction and fatal overdose~\cite{rana2025beyond, singh2019should, wen2022disentangled}, and illicit drug traffickers increasingly exploit digital platforms to evade detection and reach consumers directly~\cite{lavorgna2014internet, lavorgna2016use, broseus2016studying}. 
On the demand side, individuals susceptible to substance dependency often remain unidentified until addiction has taken hold, particularly in self-medication scenarios where individuals may not recognize their own misuse~\cite{matos2020opioids, zhang2024diet, chavez2020nutritional}.
(ii) Real-world drug-related data exhibits \textbf{structural complexity} with heterogeneous and higher-order interactions that simple homogeneous graphs may fail to model adequately.
For example, in social networks, illicit drug trafficking activities often involve group-wise interactions among multiple users, playing various roles including sellers, buyers, and discussants~\cite{hughes2017social, broseus2016studying, ma202586}.
(iii) The \textbf{publicly available, annotated datasets remain scarce} due to \textbf{privacy regulations and the need for domain expertise in labeling}.
Annotation of drug-related data is particularly labor-intensive, as it requires clinical experts to identify complex medical phenomena~\cite{goel2023llms, wei2018clinical} or legal expertise to distinguish illicit activities that vary across jurisdictions~\cite{hu2021detection, hu2021identifying}.

To this end, we introduce \textbf{\textsc{\modelname}}, the first comprehensive \textbf{OP}ioid \textbf{Bench}mark that covers five datasets to evaluate graph learning methods on real-world scenarios to combat the opioid crisis systematically. 
Specifically, to tackle the first challenge, \modelname covers three critical application domains in both supply and demand sides of the drug ecosystem, namely 
(1) \textit{Opioid Overdose Detection}, which identifies patients at risk of opioid overdose from healthcare claims data; 
(2) \textit{Online Illicit Drug Trafficking Detection} that uncover illicit drug-related activities on digital platforms; and 
(3) \textit{Drug Misuse Prediction} to leverage nutritional patterns as novel biomarkers for substance dependency.
To address the second challenge, \modelname incorporates diverse graph structures, including heterogeneous graphs, multi-relation graphs, and hypergraphs to capture the intricate dependencies and rich semantic information characterizing the opioid crisis.
To address the third challenge, we collaborate with domain experts and authoritative institutions to curate and annotate datasets while adhering to strict privacy and ethical guidelines.
We release \modelname following standardized protocols with predefined data splits, evaluation metrics, and reproducible baselines to facilitate fair and systematic comparison.
Through comprehensive experiments, we analyze the strengths and limitations of existing graph learning methods, providing actionable insights for future research in combating the opioid crisis.
Our main contributions are summarized as follows:
\begin{itemize}[leftmargin=*]
    \item \textbf{First Comprehensive Benchmark.} \textsc{\modelname} is the first comprehensive graph benchmark that enables fair and systematic comparison among graph learning methods by standardized experiment settings across five benchmark datasets that mimic the real-world scenarios to combat the severe opioid crisis. 
    \item \textbf{Systematic Empirical Analysis.} 
    We conduct comprehensive experiments across heterogeneous graphs, multi-relational graphs, and hypergraphs, revealing actionable insights into the strengths and limitations of existing graph learning methods for combating the opioid crisis.
    \item \textbf{Open-Sourced Benchmark Library.} We release \textsc{\modelname} as an easy-to-use open-sourced benchmark library to support further research and evaluation for drug-related downstream tasks. 
    With our library, researchers and practitioners can seamlessly evaluate 
    their algorithms on datasets, including but not limited to graphs, heterogeneous graphs, and hypergraphs. 
\end{itemize}

\begin{figure*}[htbp]
    \centering
    \includegraphics[width=0.98\linewidth]{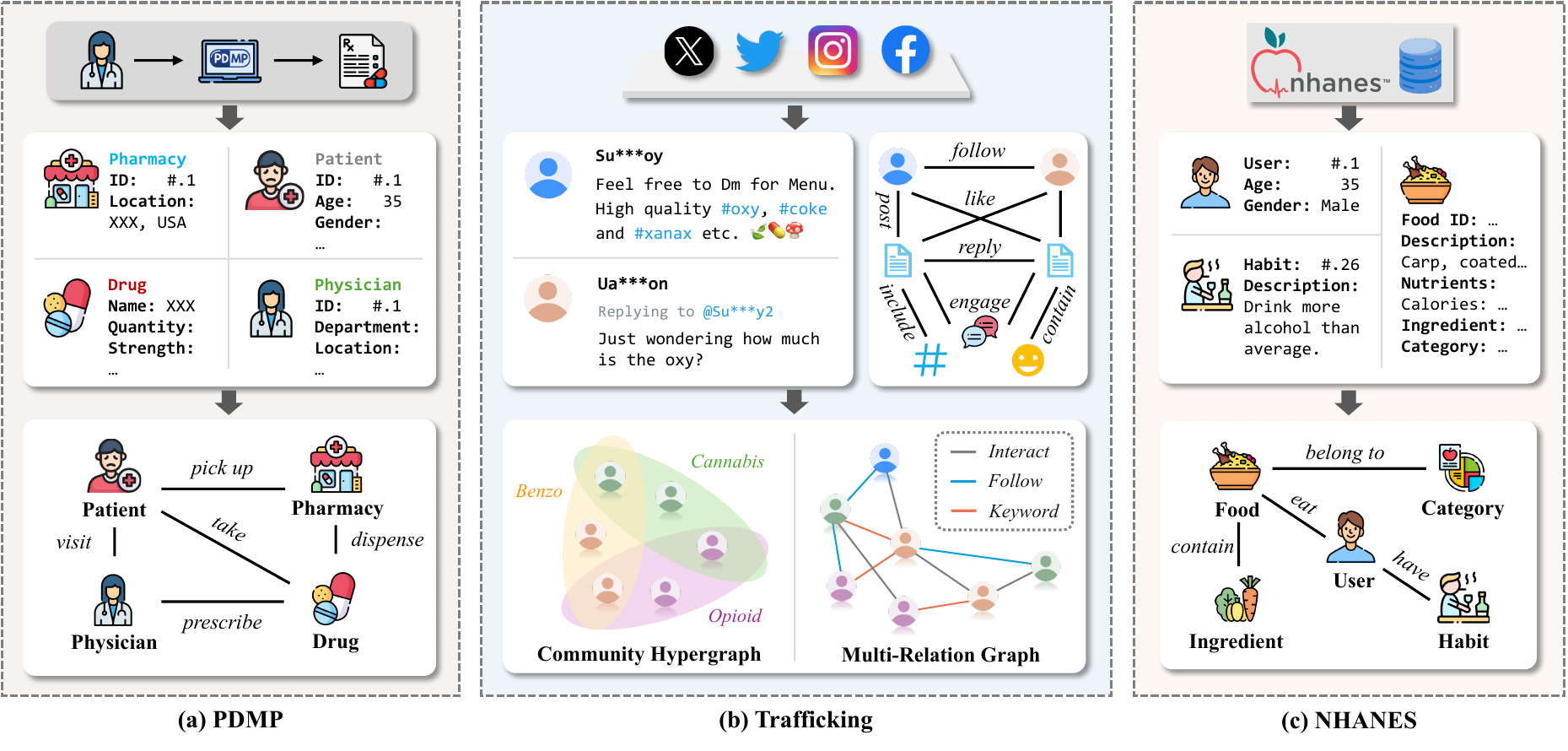}
    \caption{
        Dataset construction process for applications in \modelname. 
        (a) \textbf{\textsc{Opioid Overdose Detection}}: We leverage the PDMP dataset from the State of Ohio to construct a heterogeneous graph with four node types to detect patients at risk of opioid overdose.
        (b) \textbf{\textsc{Illicit Online Drug Trafficking}}: We collect metadata from X/Twitter and construct a hypergraph and a multi-relation graph to capture group-wise interactions and diverse relation types among users for community detection and role classification tasks.
        (c) \textbf{\textsc{Opioid Misuse Detection}}: We employ the NHANES data to construct a heterogeneous graph over five node types to predict opioid misuse through dietary patterns.
    }\label{fig: data}
\end{figure*}

\section{Backgrounds}
\subsection{Preliminary}
In this section, we provide preliminary definitions for graph, heterogeneous graph, and hypergraph, which are the fundamental data structures in our benchmark.

\begin{definition} (Graph.)
A graph is denoted as $\gG = \{\gV, \gE, \gX\}$, comprising set of nodes $\gV$ and edges $\gE$. Here, $\gX$ represents a set of node (and edge) attribute features. 
\end{definition}
\begin{definition} (Heterogeneous Graph.)
A heterogeneous graph (HG) is defined as $\gG_\text{hg} = (\gV, \gE, \gX, \gA, \gR)$, where  $\gV$ is a node set with node type set $\gA$ and  $\gE$ is the edge set with relation type set $\gR$.  
Here, $\gX = \{\rmX^A | A\in\gA \}$ represents a set of the node attribute feature matrix $\rmX^A \in \mathbb{R}^{N_A\times d_A}$ with size $N_A$ and dimension $d_A$.
\end{definition}
\begin{definition} (Hypergraph.) 
Given a hypergraph $\hp=\left (\mathcal{V}, \mathcal{E}, \mathcal{X}\right )$, $\mathcal{V}$ is the set of nodes 
with size $N$, $\mathcal{E}$ is the set of hyperedges with size $M$, and $\mathcal{X}$ is the attribute feature set. 
Each hyperedge represents a higher-order interaction among a set of nodes, i.e., $|e| \ge 2, e \in \gE$. 
\end{definition}
\subsection{Related Works}
With the excellent capability of modeling inherent relationships among entities, graph learning techniques \cite{ zhao2021multi, ju2022adaptive,wang2025graph, ju2022grape, wang2024towards, qian2022co, zhao2023self, wang2024gft} have been widely adopted in various applications including combating the opioid crisis~\cite{qian2021distilling, dong2023integrated, wen2022disentangled, rosman2023detecting, ma2023hypergraph, ma2025llm}. 
Early works have explored graph-based methods for opioid overdose prediction, leveraging the relational structure among patients, healthcare providers, and prescription records to identify individuals at elevated risk.
For example, LIGHTED~\cite{dong2023integrated} introduces a deep learning framework that integrates Long Short-Term Memory networks and Graph Neural Networks to predict patients' overdose risk by jointly modeling temporal patterns in prescription histories and spatial dependencies in patient-provider networks.
Similarly, DDHGNN~\cite{wen2022disentangled} proposes a disentangled dynamic heterogeneous graph neural network for opioid overdose prediction, which explicitly separates medical history, social environment, and provider behavior factors within a temporal graph structure to disentangle the complex interplay of clinical and socio-behavioral determinants.
NPHGS~\cite{rosman2023detecting} develops a non-parametric heterogeneous graph scan method to detect anomalous pill mill networks of prescribers and dispensers engaging in suspicious coordinated activity, enabling public health authorities to identify fraudulent prescription patterns at scale.
More recently, HyGCL-DC~\cite{ma2023hypergraph} presents a hypergraph contrastive learning framework designed to capture high-order interactions among drug sellers and drug buyers for online drug community detection, where hyperedges naturally represent group-wise co-occurrence patterns that pairwise graphs fail to adequately model.
With the advancement of large language models \cite{ye2025llms4all, li2026longda, ma2025autodata, ma2025psyscam, yuan2024mora, zhang2025agentrouter}, LLM-HetGDT~\cite{ma2025llm} designs a heterogeneous graph neural network framework that leverages large language models to oversample positive samples for online drug trafficking detection, effectively addressing the inherent class imbalance that characterizes real-world opioid crisis datasets.
Despite the significant progress in graph learning approaches to address the opioid epidemic, there remains a lack of comprehensive benchmarks tailored to systematically evaluate and analyze graph-based methods in opioid crisis scenarios.
This gap motivates our work to introduce \modelname, the first benchmark specifically designed to evaluate and analyze graph-based methods in opioid crisis applications, featuring heterogeneous graphs and hypergraphs that capture the complex relational structures inherent in public health data.

\section{Overview of \modelname}
\textsc{\modelname} fills a critical gap by aligning graph learning evaluation with the \textbf{real-world challenges} at the heart of \textbf{combating the opioid crisis}.
In this section, we provide an overview of the design and features of \textsc{\modelname}.

\subsection{Core Principles}
Our benchmark is designed following four core principles, ensuring both alignment with public health challenges and rigor in graph-based method evaluation.

\subsubsection{Alignment with Real-World Opioid Crisis Scenarios.}
Unlike existing graph benchmarks that mainly rely on bibliographic or molecular datasets~\cite{hu2020open, dwivedi2022long, li2023gslb}, \modelname is rooted in the urgent societal challenge of the opioid crisis, and the curated datasets mimic the complexity of real-world opioid-related environments.
Our benchmark covers three drivers in the crisis lifecycle. 
(i) \textbf{Prescription Opioid Supply}, where we detect and alert patients for opioid overdoses at risk from healthcare claims data, 
(ii) \textbf{Illicit Opioid Supply} that identifies illicit drug trafficking communities and roles to disrupt digital illegal supply chains, and 
(iii) \textbf{Individual Risk Indicators} leveraging nutritional patterns as latent biomarkers for early warning of opioid misuse.
This principle ensures that models evaluated on \modelname are directly relevant to high-stakes applications with tangible public health impact.

\subsubsection{High-Order and Heterogeneous Structural Modeling.}
Real-world opioid crisis data inherently involves complex, non-pairwise interactions and diverse entity types that standard homogeneous graphs fail to capture effectively~\cite{zhang2019heterogeneous,wang2025beyond,ma2023hypergraph,qian2021distilling, ma2025adaptive}. 
To address this limitation, \modelname explicitly focuses on high-order and heterogeneous structures.
(i) \textbf{Hypergraphs}. We use hyperedges to represent higher-order and group-wise interactions, yielding hypergraphs that capture complex relational information beyond pairwise interactions.
For instance, in social networks, an illicit drug trafficking community may involve multiple users collaborating, which can be naturally modeled as hyperedges.
(ii) \textbf{Heterogeneous Graphs}. We utilize heterogeneous graphs to represent the complex opioid-related systems, which is crucial for modeling the informative interactions among different types of entities and interactions.
For example, in clinical data, the interactions among patients, drugs, physician and pharmacies are inherently heterogeneous, and we define different types of nodes and edges to capture the rich semantics of these relationships to ensure a comprehensive representation of the ecosystem for modeling and analysis.
By introducing high-order and heterogeneous structures, \modelname challenges graph learning methods to handle the intricate, complex relationships and rich semantic information characterizing the opioid crisis.

\subsubsection{Data Domain Coverage.}
Recognizing that combating the opioid crisis is a multidimensional problem, \modelname spans three distinct data domains to evaluate its capabilities across diverse contexts.
(i) \textit{Healthcare}. We use Prescription Drug Monitoring Program (PDMP) data, which captures structured healthcare claims, including patient demographics, prescription records, and provider information.
(ii) \textit{Social Network}. We leverage X/Twitter data, which contains unstructured text, user profiles, and complex user interactions, to model illicit drug trafficking networks.
(iii) \textit{Nutrition}. We use data from the National Health and Nutrition Examination Survey (NHANES)~\cite{nhance}, which captures lifestyle and dietary habits, to model the correlations between dietary patterns and risk of opioid misuse.
This multi-domain coverage challenges models to adapt to varying data distributions, feature types, and structural properties.

\subsubsection{Usability and Reproducibility.}
To foster broad adoption and fair comparison, \modelname adheres to rigorous evaluation standards. 
We provide standard data splits to prevent data leakage and ensure realistic evaluation.
Furthermore, \modelname is integrated with a unified codebase that supports standard metrics, e.g., Accuracy, F1, GMeans, and AUC, and seamless compatibility with popular graph learning libraries, facilitating reproducible research and rapid benchmarking of new methods.
Moreover, this benchmark will be actively maintained and updated to incorporate emerging datasets and tasks relevant to the evolving opioid crisis.

\begin{table*}[htbp]
    \centering
    \caption{Statistics of benchmark \modelname. BC, ML, and MC denote binary,  multi-label, and multi-class classification, respectively. Het., MR. and Hp. denote heterogeneous graph, multi-relation graph, and hypergraph, respectively.
    }\label{tab: statistic} 
    \setlength{\tabcolsep}{10pt}
    \renewcommand{\arraystretch}{0.85} 
    \resizebox{\linewidth}{!}{
    \begin{tabular}{lccccccc}
        \toprule Dataset                                      & Domain         & Type                 & $|\gV|$  & $|\gE|$  & Task & $|\gY|$ & Class Ratio             \\
        \midrule\midrule
        \textsc{Pdmp-OD-Det} & Healthcare & Het. & 54,318 & 321,437 & BC & 2 & 1.0 / 1.0\\
        \textsc{X-HyDrug-Comm}                          & Social Network &  Hp.     & 2,936    & 33,893    & ML   & 6          & 1.0/1.4/2.1/1.1/0.3/4.1 \\
        \textsc{X-HyDrug-Role}                          & Social Network &  Hp.     & 2,936    & 33,893     & MC   & 4          & 1.0 / 0.8 / 4.0 / 1.2   \\
        \textsc{X-MRDrug-Role}                          & Social Network &  MR. & 27,945   &     436,385         &  MC    & 4          & 1.0 / 1.2 / 4.4 / 56.3  \\
        \textsc{NHANES-Diet}                             & Dietary        &  Het.         & 13,742   &    445,468     & BC & 2 & 1.0/1.0            \\
        \bottomrule
    \end{tabular}
    }
\end{table*}

\subsection{Baseline Methods}
Our benchmark framework supports all graph learning methods seamlessly.
We implement classic baseline methods in the literature, which can be categorized as 
(i) Feature-based methods:  Multi-layer Perceptrons (MLPs), Oversampling~\cite{zheng2015oversampling}, and SMOTE~\cite{chawla2002smote};
(ii) Graph Neural Networks (GNNs): GCN~\cite{kipf2017semisupervisedclassificationgraphconvolutional}, GAT~\cite{velivckovic2017graph}, Sage~\cite{hamilton2017inductive}, and GraphSMOTE~\cite{zhao2021graphsmote};
(iii) Heterogeneous Graph Neural Networks (HetGNNs): HGMAE~\cite{tian2023heterogeneous}, HAN~\cite{wang2019heterogeneous}, and HGT~\cite{hu2020heterogeneous};
(iv) Multi-Relation Graph Neural Networks (MRGNNs): R-GCN~\cite{schlichtkrull2018modeling}, GraphENS~\cite{park2021graphens}, and AD-GSMOTE~\cite{qian2025adaptive};
(v) Hypergraph Neural Networks (HyGNNs): HGNN~\cite{feng2019hypergraph}, HNHN~\cite{dong2020hnhn}, HCHA~\cite{bai2021hypergraph}, AllSet~\cite{chien2021you}, and ED-HNN~\cite{wang2022equivariant}.
Implementation details for each method are provided in Appendix~\ref{app: baseline}.

\subsection{Evaluation Protocols}
To ensure a rigorous and fair evaluation, we carefully design the evaluation protocols in \modelname as follows.

\noindent\textbf{Evaluation Metrics.} We adopt standard evaluation metrics to comprehensively assess model performance across different aspects, such as overall correctness (Accuracy), balance between precision and recall (F1-score), area under the correctness curve (AUC), and imbalanced performance (GMeans).

\noindent\textbf{Data Splits.} To prevent data leakage and ensure reproducible evaluation, we provide standard data splits for training, validation, and testing, with multiple ratios, i.e., 10\%, 20\%, 50\% training data, 10\% for validation, and the rest for testing to assess model robustness under varying data availability scenarios.
For each data split ratio, we provide five different random splits and report the average performance with standard deviation across the 5 splits to ensure a robust and reliable evaluation of model performance.

\noindent\textbf{Hyperparameter Tuning.} To find the optimal hyperparameters for each baseline method, we perform a grid search over a predefined hyperparameter space, including learning rate \{0.1, 0.01, 0.001, 0.0001\}, hidden dimension \{64, 128, 256, 512\}, dropout rate \{0.0, 0.3, 0.5\}, and weight decay \{0, 0.0001, 0.0005\}.
Note that for a fair comparison, we fix the number of layers to 2 for all baseline methods, and we tune the hyperparameters based on the validation set performance.
Moreover, each method is attached with a two-layer MLP as a classifier for downstream tasks.
We report the best hyperparameters for each method and dataset in Appendix Table~\ref{tab: parameters}.

\noindent\textbf{Runtime Environment.} All experiments are conducted using Nvidia A40 GPUs with 48 GB GDDR6 memories with PyTorch 2.9.1 and PyG 2.7.0 to ensure consistent performance measurements.

\section{\modelname Datasets}

In this section, we present the details of \modelname, which consists of three critical domains in the opioid crisis lifecycle with five datasets curated from real-world data sources.
For each domain, we describe its real-world significance in combating the opioid crisis, our contributions, data construction methods, and experiment results.
The data statistics are summarized in Table~\ref{tab: statistic}.

\begin{table*}[htbp]
    \centering
    \caption{The performance comparison among \textsc{Pdmp-OD-Det}.
    The best performance is bolded and runner-ups are underlined.
        }
    \renewcommand{\arraystretch}{0.9} 
    \label{tab: pdmp-od-det}
    \resizebox{\linewidth}{!}
    {
    \begin{tabular}{lccccccccc}
       \toprule
       \multirow{2}{*}{Model} &  \multicolumn{3}{c}{Train-10\% Valid-10\% Test-80\%} & \multicolumn{3}{c}{Train-20\% Valid-10\% Test-70\%} & \multicolumn{3}{c}{Train-50\% Valid-10\% Test-40\%} \\
       \cmidrule(lr){2-4} \cmidrule(lr){5-7} \cmidrule(lr){8-10}
       & AUC & F1-Macro & F1-Micro & AUC & F1-Macro & F1-Micro & AUC & F1-Macro & F1-Micro \\ 
       \midrule\midrule
       MLP & 72.57{\tiny $\pm$6.86}  & 68.77{\tiny $\pm$4.97} & 67.06{\tiny $\pm$6.54} & 77.02{\tiny $\pm$0.69} & 71.12{\tiny $\pm$6.32} & 72.40{\tiny $\pm$5.79} &  75.41{\tiny $\pm$2.67} & 60.52{\tiny $\pm$8.59} & 66.18{\tiny $\pm$8.59} \\
       GCN~\cite{kipf2017semisupervisedclassificationgraphconvolutional} & 60.96{\tiny $\pm$0.35} &52.95{\tiny $\pm$6.58} & 57.86{\tiny $\pm$0.27} & 68.58{\tiny $\pm$0.26} & 57.58{\tiny $\pm$8.47} & 60.20{\tiny $\pm$4.35} & 70.91{\tiny $\pm$0.55} & 66.99{\tiny $\pm$8.01} & 67.13{\tiny $\pm$3.35}\\
       GAT~\cite{velivckovic2017graph} & 
       61.76{\tiny $\pm$4.48} & 53.94{\tiny $\pm$ 7.67} & 56.70{\tiny $\pm$3.11} & 70.75{\tiny $\pm$2.49} & 59.10{\tiny $\pm$10.36} & 61.77{\tiny $\pm$5.60} & 75.78{\tiny $\pm$2.85} & 69.64{\tiny $\pm$6.43} & 70.82{\tiny $\pm$2.20}\\ 
       R-GCN~\cite{schlichtkrull2018modeling} & 63.37{\tiny $\pm$3.95} & 55.85{\tiny $\pm$10.78} & 58.99{\tiny $\pm$4.51}& 70.54{\tiny $\pm$3.99} & 59.79{\tiny $\pm$12.91} & 63.14{\tiny $\pm$6.37} & 73.94{\tiny $\pm$0.25} & 67.93{\tiny $\pm$ 0.33} & 68.03{\tiny $\pm$0.34}\\
       HAN~\cite{wang2019heterogeneous} & \underline{79.14{\tiny $\pm$0.45}} & \textbf{75.78{\tiny $\pm$4.92}} & \textbf{74.82{\tiny $\pm$4.73}} & \textbf{84.11{\tiny $\pm$4.82}} & \textbf{76.45{\tiny $\pm$11.09}} & \textbf{79.44{\tiny $\pm$3.05}} & \textbf{87.06{\tiny $\pm$4.11}} & \underline{78.02{\tiny $\pm$9.22}} & \textbf{82.41{\tiny $\pm$2.54}}\\ 
       HGT~\cite{hu2020heterogeneous} & \textbf{79.81{\tiny $\pm$0.88}} & \underline{70.97{\tiny $\pm$7.91}} & \underline{72.67{\tiny $\pm$4.54}} & \underline{83.21{\tiny $\pm$6.67}} & \underline{74.22{\tiny $\pm$13.05}} & \underline{77.66{\tiny $\pm$6.55}} & \underline{85.21{\tiny $\pm$6.67}} & \textbf{79.22{\tiny $\pm$13.05}} & \underline{81.66{\tiny $\pm$6.55}}\\ 
       HGMAE~\cite{tian2023heterogeneous} & 62.89{\tiny $\pm$3.38} & 51.53{\tiny $\pm$10.02} & 56.93{\tiny $\pm$2.32} & 66.77{\tiny $\pm$0.53} & 62.74{\tiny $\pm$0.62} & 64.03{\tiny $\pm$0.62} & 70.43{\tiny $\pm$0.68} & 67.58{\tiny $\pm$0.51} & 69.70{\tiny $\pm$0.26}\\
       
        \bottomrule
    \end{tabular}
    }
\end{table*}

\begin{table*}[htbp]
    \centering
    \caption{The performance comparison among \textsc{X-HyDrug-Comm}.
    The best performance is bolded, and runner-ups are underlined.
    }
    \label{tab: X-hydrug-comm}
    \renewcommand{\arraystretch}{0.9} 
    \resizebox{\linewidth}{!}
    {
    \begin{tabular}{lccccccccc}
       \toprule
       \multirow{2}{*}{Model} &  \multicolumn{3}{c}{Train-10\% Valid-10\% Test-80\%} & \multicolumn{3}{c}{Train-20\% Valid-10\% Test-70\%} & \multicolumn{3}{c}{Train-50\% Valid-10\% Test-40\%} \\
       \cmidrule(lr){2-4} \cmidrule(lr){5-7} \cmidrule(lr){8-10}
       & Acc. & F1-Macro & F1-Micro & Acc. & F1-Macro & F1-Micro & Acc. & F1-Macro & F1-Micro\\ 
       \midrule\midrule
       MLP   & \underline{82.63}{\tiny $\pm$0.68} & 47.55{\tiny $\pm$4.27} & 70.55{\tiny $\pm$2.11} & \underline{84.05}{\tiny $\pm$0.43} & \underline{58.48}{\tiny $\pm$4.29} & \underline{74.60}{\tiny $\pm$1.07} & 84.59{\tiny $\pm$0.38} & 56.13{\tiny $\pm$3.44} & 74.96{\tiny $\pm$0.97} \\
       GCN~\cite{kipf2017semisupervisedclassificationgraphconvolutional} & 78.68{\tiny $\pm$0.23} & 39.09{\tiny $\pm$0.97} & 63.58{\tiny $\pm$0.76} & 80.14{\tiny $\pm$0.24} & 41.37{\tiny $\pm$1.84} & 65.36{\tiny $\pm$0.28} & 80.66{\tiny $\pm$0.33} & 45.39{\tiny $\pm$2.32} & 66.67{\tiny $\pm$1.19} \\
       GAT~\cite{velivckovic2017graph} & 79.42{\tiny $\pm$0.51} & 42.21{\tiny $\pm$2.28} & 64.72{\tiny $\pm$0.78} & 80.44{\tiny $\pm$0.59} & 47.54{\tiny $\pm$2.75} & 66.68{\tiny $\pm$1.75} & 82.10{\tiny $\pm$0.54} & 55.12{\tiny $\pm$1.68} & 71.08{\tiny $\pm$1.19} \\
       HGNN~\cite{feng2019hypergraph}  & 80.70{\tiny $\pm$0.16} & 45.40{\tiny $\pm$1.40} & 65.84{\tiny $\pm$0.20} & 82.31{\tiny $\pm$0.32} & 49.96{\tiny $\pm$1.66} & 69.96{\tiny $\pm$0.93} & 82.49{\tiny $\pm$2.88} & 50.97{\tiny $\pm$2.88} & 69.77{\tiny $\pm$0.77} \\
       HNHN~\cite{dong2020hnhn}  & 82.55{\tiny $\pm$0.65} & \underline{54.64}{\tiny $\pm$1.78} & \underline{71.80}{\tiny $\pm$1.37} & 83.89{\tiny $\pm$0.48} & 57.60{\tiny $\pm$4.97} & 74.26{\tiny $\pm$1.02} & 85.05{\tiny $\pm$0.86} & 58.07{\tiny $\pm$3.28} & 76.01{\tiny $\pm$1.71} \\
       HCHA~\cite{bai2021hypergraph}  & 82.20{\tiny $\pm$0.31} & 53.72{\tiny $\pm$1.58} & 70.71{\tiny $\pm$0.74} & 83.12{\tiny $\pm$0.27} & 57.94{\tiny $\pm$1.52} & 72.41{\tiny $\pm$0.56} & 84.28{\tiny $\pm$0.48} & \underline{64.10}{\tiny $\pm$1.67} & 74.82{\tiny $\pm$1.04} \\
       AllSet~\cite{chien2021you} & 80.69{\tiny $\pm$1.35} & 41.70{\tiny $\pm$6.98} & 66.68{\tiny $\pm$3.76} & 83.06{\tiny $\pm$0.52} & 52.40{\tiny $\pm$1.92} & 72.16{\tiny $\pm$0.79} & \underline{85.35}{\tiny $\pm$0.46} & 63.94{\tiny $\pm$2.33} & \underline{77.00}{\tiny $\pm$1.05} \\
       ED-HNN~\cite{wang2022equivariant} & \textbf{83.77}{\tiny $\pm$0.36} & \textbf{57.62}{\tiny $\pm$2.79} & \textbf{74.57}{\tiny $\pm$0.37} & \textbf{85.29}{\tiny $\pm$0.32} & \textbf{65.10}{\tiny $\pm$4.80} & \textbf{76.66}{\tiny $\pm$0.60} & \textbf{86.59}{\tiny $\pm$0.42} & \textbf{73.39}{\tiny $\pm$1.24} & \textbf{79.08}{\tiny $\pm$0.76} \\
        \bottomrule
    \end{tabular}
    }
\end{table*}

\subsection{Healthcare: Opioid Overdose}
\textbf{Why this matters.}
The opioid crisis represents one of the deadliest public health emergencies in modern history, claiming over 80,000 lives annually in the United States alone~\cite{carrillo2022lancet}.
As opioids remain among the most widely prescribed medications for pain management, their addictive properties place patients at significant risk of dependency and fatal overdose~\cite{mehta2006acute, mcquay1999opioids}.
Although opioids require physician prescriptions, overdose deaths continue to rise, suggesting that certain prescribing patterns may inadvertently contribute to patient harm~\cite{judd2023opioid, chen2019prevention}.
Critically, opioid risk is not determined by patient characteristics alone.
It emerges from complex interactions among patients, prescribers, pharmacies, and medications.
A patient receiving prescriptions from multiple providers, or a pharmacy dispensing unusually high volumes of controlled substances, may signal an elevated risk that isolated patient records cannot reveal.
Identifying such high-risk patterns before adverse outcomes occur is essential to effective prevention, yet manual evaluation by healthcare professionals proves inadequate at scale due to limited bandwidth and the sheer volume of prescription data.
Graph learning methods are uniquely suited to this challenge, as they can model the relational structure among patients, medications, and providers embedded in healthcare claims data to detect risk signals invisible to traditional approaches.

\noindent\textbf{Our Contribution.} 
We frame this as a node classification task that aims to predict whether the patient in a healthcare network is at high risk for opioid overdose.
To evaluate graph learning methods on this task, we release \textsc{Pdmp-OD-Det}, a heterogeneous graph dataset built from Ohio's PDMP data.
PDMPs are state-run databases that track every controlled substance prescription filled in the state, providing a comprehensive view of prescribing and dispensing patterns.

\subsubsection{Dataset Construction and Annotation}
We obtain the PDMP data from the State of Ohio Board of Pharmacy spanning the year 2016, which contains 1,395,958 medical prescribing and dispensing records throughout the year.
We first remove incomplete records, e.g., missing demographic information, etc., leading to a total of 1,395,434 records over 61,479 patients.
Afterward, we construct a heterogeneous graph \het that contains four node types and five edge types, as illustrated in Figure~\ref{fig: data}(a).
We follow existing works~\cite{lossio2022opioid2mme, adams2025standardizing} to annotate the patients with binary labels, where the positive class represents patients facing high opioid overdose risk, and the negative class represents patients at low risk.
Particularly, we employ the Morphine Milligram Equivalent (MME)~\cite{lossio2022opioid2mme} stressed by the CDC as a standard indicator for opioid overdose risk.
Following the CDC-recommended daily MME threshold~\cite{dowell2022cdc}, patients with an average daily MME above 90 are labeled as high-risk patients, while patients falling below this threshold are identified as low-risk patients.
As a result, we obtain \textsc{Pdmp-OD-Det} dataset with 54,318 nodes and 321,437 edges.
More details about data construction and annotation are provided in Appendix~\ref{app: pdmp}, and data statistics are summarized in Appendix Table~\ref{tab:pdmp_stat}.

\subsubsection{Performance Comparison for Opioid Overdose Detection.}
Table~\ref{tab: pdmp-od-det} reports the performance comparison among baseline methods on the \textsc{Pdmp-OD-Det} dataset.
According to the table, we observe that: 
(i) Feature-based method, i.e., MLPs, outperform GNNs, including GCN and GAT, in several settings, indicating that message passing on projected homogeneous graphs fails to capture the rich heterogeneous semantics and may even introduce noise.
(ii) HetGNNs, including HAN and HGT, significantly outperform GNNs and MLPs across all training ratios and evaluation metrics, demonstrating that explicitly modeling the heterogeneous relationships among patients, physicians, pharmacies, and drugs is essential for accurate opioid overdose risk prediction.
(iii) HAN and HGT exhibit competitive performance, with HAN achieving the best results in most metrics while HGT leads in AUC at 10\% training ratio and F1-Macro at 50\% training ratio.
This suggests that both hierarchical attention and heterogeneous graph transformers effectively capture the complex semantics of healthcare networks, though their relative strengths vary with data availability.

\begin{table*}[htbp]
    \centering
    \caption{The performance comparison among \textsc{X-HyDrug-Role}.
    The best performance is bolded and runner-ups are underlined.
    }
    \label{tab: X-hydrug-role}
    \renewcommand{\arraystretch}{0.9} 
    \resizebox{\linewidth}{!}
    {
    \begin{tabular}{lccccccccc}
       \toprule
       \multirow{2}{*}{Model} &  \multicolumn{3}{c}{Train-10\% Valid-10\% Test-80\%} & \multicolumn{3}{c}{Train-20\% Valid-10\% Test-70\%} & \multicolumn{3}{c}{Train-50\% Valid-10\% Test-40\%} \\
       \cmidrule(lr){2-4} \cmidrule(lr){5-7} \cmidrule(lr){8-10}
       & Acc. & F1-Macro & F1-Micro & Acc. & F1-Macro & F1-Micro & Acc. & F1-Macro & F1-Micro\\ 
       \midrule\midrule
       MLP      & 67.46{\tiny $\pm$0.53} & 45.25{\tiny $\pm$2.37} & 67.46{\tiny $\pm$0.53} & 68.35{\tiny $\pm$1.03} & \underline{47.60}{\tiny $\pm$1.40} & 68.35{\tiny $\pm$1.03} & 69.58{\tiny $\pm$1.75} & 48.89{\tiny $\pm$2.25} & 69.58{\tiny $\pm$1.75} \\
       GCN~\cite{kipf2017semisupervisedclassificationgraphconvolutional}    & 64.87{\tiny $\pm$0.51} & 40.55{\tiny $\pm$0.66} & 64.87{\tiny $\pm$0.51} & 65.20{\tiny $\pm$0.78} & 39.36{\tiny $\pm$0.76} & 65.20{\tiny $\pm$0.78} & 66.08{\tiny $\pm$1.12} & 44.49{\tiny $\pm$2.06} & 66.08{\tiny $\pm$1.12} \\
       GAT~\cite{velivckovic2017graph}    & 65.57{\tiny $\pm$0.81} & 40.37{\tiny $\pm$1.44} & 65.57{\tiny $\pm$0.81} & 66.42{\tiny $\pm$1.28} & 41.26{\tiny $\pm$1.55} & 66.42{\tiny $\pm$1.28} & 67.35{\tiny $\pm$2.04} & 41.49{\tiny $\pm$1.65} & 67.35{\tiny $\pm$2.04} \\
       HGNN~\cite{feng2019hypergraph}     & \underline{67.69}{\tiny $\pm$0.21} & 43.45{\tiny $\pm$2.00} & \underline{67.69}{\tiny $\pm$0.21} & \underline{68.64}{\tiny $\pm$0.90} & 46.06{\tiny $\pm$1.86} & \underline{68.64}{\tiny $\pm$0.90} & 69.38{\tiny $\pm$0.92} & 49.02{\tiny $\pm$2.78} & 69.38{\tiny $\pm$0.92} \\
       HNHN~\cite{dong2020hnhn}   & 66.89{\tiny $\pm$0.63} & 40.83{\tiny $\pm$1.43} & 66.89{\tiny $\pm$0.63} & 67.90{\tiny $\pm$0.86} & 41.11{\tiny $\pm$1.49} & 67.90{\tiny $\pm$0.86} & 68.94{\tiny $\pm$1.55} & 45.40{\tiny $\pm$3.88} & 68.94{\tiny $\pm$1.55} \\
       HCHA~\cite{bai2021hypergraph}     & 67.28{\tiny $\pm$0.53} & \textbf{45.91}{\tiny $\pm$2.22} & 67.28{\tiny $\pm$0.53} & 68.09{\tiny $\pm$0.85} & \textbf{48.34}{\tiny $\pm$1.96} & 68.09{\tiny $\pm$0.85} & 68.83{\tiny $\pm$1.21} & 48.98{\tiny $\pm$1.24} & 68.83{\tiny $\pm$1.21} \\
       AllSet~\cite{chien2021you}   & 67.09{\tiny $\pm$0.49} & 42.71{\tiny $\pm$2.90} & 67.09{\tiny $\pm$0.49} & 68.09{\tiny $\pm$0.71} & 43.20{\tiny $\pm$2.49} & 68.09{\tiny $\pm$0.71} & \underline{69.92}{\tiny $\pm$0.40} & \textbf{52.23}{\tiny $\pm$3.95} & \underline{69.92}{\tiny $\pm$0.40} \\
       ED-HNN~\cite{wang2022equivariant}   & \textbf{68.55}{\tiny $\pm$0.62} & \underline{45.58}{\tiny $\pm$3.99} & \textbf{68.55}{\tiny $\pm$0.62} & \textbf{69.40}{\tiny $\pm$1.08} & 45.12{\tiny $\pm$3.53} & \textbf{69.40}{\tiny $\pm$1.08} & \textbf{70.66}{\tiny $\pm$1.84} & \underline{49.08}{\tiny $\pm$4.52} & \textbf{70.66}{\tiny $\pm$1.84} \\
        \bottomrule
    \end{tabular}
    }
\end{table*}

\begin{table*}[htbp]
    \centering
    \caption{The performance comparison among \textsc{X-MrDrug-Role}.
    The best performance is bolded and runner-ups are underlined.
    }
    \label{tab: X-mrdrug-role}
    \renewcommand{\arraystretch}{0.9} 
    \resizebox{\linewidth}{!}
    {
    \begin{tabular}{lccccccccc}
       \toprule
       \multirow{2}{*}{Model} &  \multicolumn{3}{c}{Train-10\% Valid-10\% Test-80\%} & \multicolumn{3}{c}{Train-20\% Valid-10\% Test-70\%} & \multicolumn{3}{c}{Train-50\% Valid-10\% Test-40\%} \\
       \cmidrule(lr){2-4} \cmidrule(lr){5-7} \cmidrule(lr){8-10}
       & GMean & F1-Macro & F1-Micro & GMean & F1-Macro & F1-Micro & GMean & F1-Macro & F1-Micro\\ 
       \midrule\midrule
       MLP   & 40.53{\tiny $\pm$12.80}&62.38{\tiny $\pm$2.40} & 94.43{\tiny $\pm$0.11}& 52.60{\tiny $\pm$2.61}&66.43{\tiny $\pm$1.20} & 94.94{\tiny $\pm$0.17} & 59.74{\tiny $\pm$1.57}&70.30{\tiny $\pm$1.09} & 95.56{\tiny $\pm$0.29}  \\
        Oversampling~\cite{zheng2015oversampling} & 57.61{\tiny $\pm$2.11}&62.66{\tiny $\pm$0.77} & 92.40{\tiny $\pm$0.49}& 60.83{\tiny $\pm$1.35}&64.33{\tiny $\pm$0.67} & 92.59{\tiny $\pm$0.52} & \underline{65.53}{\tiny $\pm$2.53}&68.45{\tiny $\pm$1.95} & 94.10{\tiny $\pm$0.41}  \\
       SMOTE~\cite{chawla2002smote} & \underline{57.98}{\tiny $\pm$3.68}&62.68{\tiny $\pm$1.69} & 92.50{\tiny $\pm$1.09}& 60.92{\tiny $\pm$0.99}&65.52{\tiny $\pm$0.63} & 93.49{\tiny $\pm$0.09} & 65.52{\tiny $\pm$2.78}&69.10{\tiny $\pm$1.53} & 94.31{\tiny $\pm$0.22}  \\

       GCN~\cite{kipf2017semisupervisedclassificationgraphconvolutional} & 33.84{\tiny $\pm$20.01}&60.31{\tiny $\pm$3.59} & 94.48{\tiny $\pm$0.10}& 52.03{\tiny $\pm$4.45}&64.38{\tiny $\pm$1.64} & 94.75{\tiny $\pm$0.18} & 53.55{\tiny $\pm$3.69}&66.08{\tiny $\pm$1.38} & 95.22{\tiny $\pm$0.26}  \\
       GAT~\cite{velivckovic2017graph}  & 26.87{\tiny $\pm$16.89}&53.43{\tiny $\pm$3.39} & 91.35{\tiny $\pm$0.41}& 44.79{\tiny $\pm$8.41}&58.63{\tiny $\pm$3.33} & 92.78{\tiny $\pm$0.52} & 55.67{\tiny $\pm$3.44}&65.55{\tiny $\pm$1.73} & 94.40{\tiny $\pm$0.37}  \\
       GraphSage~\cite{hamilton2017inductive}  & 50.97{\tiny $\pm$15.62}&\underline{67.28}{\tiny $\pm$4.02} & \underline{95.32}{\tiny $\pm$0.16}& \underline{60.93}{\tiny $\pm$6.34}&\underline{70.73}{\tiny $\pm$2.13} & \underline{95.87}{\tiny $\pm$0.06} & 63.97{\tiny $\pm$2.10}&\underline{72.82}{\tiny $\pm$1.05} & \underline{96.30}{\tiny $\pm$0.16}  \\
       GraphENS~\cite{park2021graphens} & 38.84{\tiny $\pm$13.83}&60.61{\tiny $\pm$3.18} & 94.56{\tiny $\pm$0.06}& 49.71{\tiny $\pm$4.36}&63.40{\tiny $\pm$1.45} & 94.78{\tiny $\pm$0.13} & 49.39{\tiny $\pm$7.71}&64.98{\tiny $\pm$2.51} & 95.19{\tiny $\pm$0.21}  \\
       GraphSMOTE~\cite{zhao2021graphsmote}  & 38.15{\tiny $\pm$21.85}&64.93{\tiny $\pm$4.04} & 95.32{\tiny $\pm$0.13}& 55.33{\tiny $\pm$10.34}&69.61{\tiny $\pm$2.99} & 95.87{\tiny $\pm$0.07} & 61.71{\tiny $\pm$3.94}&72.66{\tiny $\pm$1.80} & 96.30{\tiny $\pm$0.20}  \\
        AD-GSMOTE~\cite{qian2025adaptive}  & \textbf{61.68}{\tiny $\pm$1.59}&\textbf{70.58}{\tiny $\pm$0.99} & \textbf{95.51}{\tiny $\pm$0.14}& \textbf{67.00}{\tiny $\pm$0.91}&\textbf{75.02}{\tiny $\pm$0.33} & \textbf{96.28}{\tiny $\pm$0.05} & \textbf{69.10}{\tiny $\pm$1.02}&\textbf{77.70}{\tiny $\pm$0.64} & \textbf{96.71}{\tiny $\pm$0.11}  \\
        \bottomrule
    \end{tabular}
    }
\end{table*}

\subsection{Social Network: Illicit Drug Trafficking}
\textbf{Why this matters.} 
As the market of illicit drugs remains immensely profitable, the crime of drug trafficking has continually evolved with modern technologies, such as social media platforms.
Recent works~\cite{qian2021distilling,ma2023hypergraph} have shown that the major social media platforms, e.g., X/Twitter, Instagram, and Facebook, have become the direct-to-customer intermediaries for illicit drug trafficking, especially for opioid substances. 
As illustrated in Fig.~\ref{fig: data}(b), illicit drug sellers leverage social media to advertise their products through posts containing street names and drug-related hashtags, which attract potential buyers and facilitate direct or indirect drug trade on digital platforms.
Consequently, these drug-related illicit behaviors among users naturally form drug trafficking networks on social media platforms, and these illegal drug-related activities have turned into a societal concern due to their catastrophic consequences, from violent crimes to public health.

\noindent\textbf{Our Contribution.}
To evaluate the performance of graph learning methods, we present three benchmark datasets for online drug trafficking detection, i.e., \textsc{X-HyDrug-Role}, \textsc{X-HyDrug-Comm}, and \textsc{X-MRDrug-Role}, where each dataset aims to tackle a specific challenge in illicit online drug trafficking detection.
\textbf{(i) Online drug trafficking community.} 
Recent works~\cite{garcia2009labor, browne2003drug} have demonstrated that drug-related users operate in tightly-knit communities to facilitate their illicit activities. 
To this end, we introduce \textsc{X-HyDrug-Comm}, a hypergraph benchmark that mimics the multi-community nature of illicit drug trafficking networks for the community detection task.
Moreover, we formulate the problem as an overlapping community detection task, as individuals involved in online drug trafficking may belong to multiple communities.
\textbf{(ii) Roles in online illicit drug trafficking.} 
In addition to community detection, we construct \textsc{X-HyDrug-Role} to capture higher-order interactions among users: drug-relevant users collaborate in groups to conduct their illicit activities, and these group-level interactions can be modeled as hyperedges in a hypergraph. 
We formulate this problem as a drug trafficking role detection task to identify the specific roles of users in illicit online drug trafficking activities, i.e., drug sellers, drug buyers, drug users, and discussion.
\textbf{(iii) Minority issue in illicit online drug trafficking.}
Illicit drug-relevant users remain a small portion of users compared to the normal users in the real-world digital platform, and this poses significant challenges for model training and detecting illicit drug-related users effectively~\cite{qian2025adaptive}.
In response, we propose a multi-relation graph, called \textsc{X-MRDrug-Role}, that mimics the imbalanced label distribution in real-world scenarios. 
Moreover, \textsc{X-MRDrug-Role} models pairwise interactions among users, as drug-relevant users also interact with one another through various types of relations, e.g., follow, mention.
These multi-relation types can be effectively captured by multi-relation graphs, in which different edge types represent distinct interactions among users.
These three datasets provide a comprehensive benchmark for evaluating graph learning methods for drug trafficking detection across various settings, including community and role detection, and different graph structures, i.e., hypergraphs and multi-relational graphs.

\begin{table*}[htbp]
    \centering
    \caption{The performance comparison among \textsc{NHANES-Diet}.
    The best performance is bolded and runner-ups are underlined.
    }
    \label{tab: nhance-diet-role}
    \renewcommand{\arraystretch}{0.91} 
    \resizebox{\linewidth}{!}
    {
    \begin{tabular}{lccccccccc}
       \toprule
       \multirow{2}{*}{Model} &  \multicolumn{3}{c}{Train-10\% Valid-10\% Test-80\%} & \multicolumn{3}{c}{Train-20\% Valid-10\% Test-70\%} & \multicolumn{3}{c}{Train-50\% Valid-10\% Test-40\%} \\
       \cmidrule(lr){2-4} \cmidrule(lr){5-7} \cmidrule(lr){8-10}
       & Acc. & F1-Macro & F1-Micro & Acc. & F1-Macro & F1-Micro & Acc. & F1-Macro & F1-Micro\\ 
       \midrule\midrule
       MLP &68.87{\tiny $\pm$ 1.27}&69.27{\tiny $\pm$ 2.30}&68.87{\tiny $\pm$ 1.27}&70.62{\tiny $\pm$ 0.62}&71.49{\tiny $\pm$ 0.90}&70.62{\tiny $\pm$ 0.62}&72.76{\tiny $\pm$ 0.89}&73.90{\tiny $\pm$ 0.97}& 72.76{\tiny $\pm$ 0.89}\\
       GCN~\cite{kipf2017semisupervisedclassificationgraphconvolutional} &71.51{\tiny $\pm$ 1.20}&73.01{\tiny $\pm$ 1.48}&71.51{\tiny $\pm$ 1.20}&73.46{\tiny $\pm$ 0.85}&75.53{\tiny $\pm$ 0.90}&73.46{\tiny $\pm$ 0.85}&74.90{\tiny $\pm$ 0.60}&77.15{\tiny $\pm$ 0.67}&74.90{\tiny $\pm$ 0.60} \\
       GAT~\cite{velivckovic2017graph} & \underline{73.74}{\tiny $\pm$ 0.74} &\underline{76.12}{\tiny $\pm$ 1.01}&\underline{73.74}{\tiny $\pm$ 0.74}&\underline{74.97}{\tiny $\pm$ 0.64}&\underline{77.36}{\tiny $\pm$ 0.66}&\underline{74.97}{\tiny $\pm$ 0.64}&75.32{\tiny $\pm$ 0.90}&77.80{\tiny $\pm$ 0.90}&75.32{\tiny $\pm$ 0.90}\\ 
       R-GCN~\cite{schlichtkrull2018modeling} &72.74{\tiny $\pm$ 1.06} &74.73{\tiny $\pm$ 1.25}&72.74{\tiny $\pm$ 1.06}&74.26{\tiny $\pm$ 0.70}&76.62{\tiny $\pm$ 0.60}&74.26{\tiny $\pm$ 0.70}&\underline{75.42}{\tiny $\pm$ 0.66}&\underline{77.96}{\tiny $\pm$ 0.58}&\underline{75.42}{\tiny $\pm$ 0.66} \\
       HAN~\cite{wang2019heterogeneous} &\textbf{74.06}{\tiny $\pm$ 0.82}&\textbf{76.27}{\tiny $\pm$ 1.07}&\textbf{74.06}{\tiny $\pm$ 0.82}&\textbf{75.10}{\tiny $\pm$ 0.55}&\textbf{77.69}{\tiny $\pm$ 0.37}&\textbf{75.10}{\tiny $\pm$ 0.55}&\textbf{75.90}{\tiny $\pm$ 0.89}&\textbf{78.46}{\tiny $\pm$ 0.84}&\textbf{75.90}{\tiny $\pm$ 0.89}\\ 
       HGT~\cite{hu2020heterogeneous} &70.58{\tiny $\pm$ 1.89}&71.27{\tiny $\pm$ 2.53}&70.58{\tiny $\pm$ 1.89}&72.38{\tiny $\pm$ 1.01}&74.03{\tiny $\pm$ 2.67}&72.38{\tiny $\pm$ 1.01}&75.41{\tiny $\pm$ 1.58}&73.33{\tiny $\pm$ 1.11}& 75.41{\tiny $\pm$ 1.58}\\
       HGMAE~\cite{tian2023heterogeneous} & 63.17{\tiny $\pm$ 0.61}&63.10{\tiny $\pm$ 0.86}&63.17{\tiny $\pm$ 0.61}&63.80{\tiny $\pm$ 0.93}&63.57{\tiny $\pm$ 1.59}&63.80{\tiny $\pm$ 0.93}&64.83{\tiny $\pm$ 0.82}&64.80{\tiny $\pm$ 0.73}&64.83{\tiny $\pm$ 0.82}\\
       
        \bottomrule
    \end{tabular}
    }
\end{table*}

\subsubsection{Dataset Construction and Annotation}
We choose X (formerly Twitter) as the social media platform to study the drug trafficking detection tasks, and leveraged the official APIs to collect posts and user profiles.
With the collected metadata, six experts are recruited to manually annotate the users based on their posts and profiles, following the annotation rules (Appendix~\ref{app: x}).
With the annotated data, we construct the heterogeneous graph \het for \textsc{X-MRDrug-Role} and the hypergraph \hp for \textsc{X-HyDrug-Comm} and \textsc{X-HyDrug-Role} following the schema as illustrated in Fig.~\ref{fig: data}(b).
Specifically, for the heterogeneous graph \het, we define three edge types over user nodes, i.e., \textit{user-interact-user}, \textit{user-follow-user}, and \textit{user-keyword-user}, where the first two edge types capture the interactions among users, and the last edge type captures the shared interests among users through their common keywords, e.g., hashtags and emojis.
Following existing works~\cite{ma2023hypergraph, wang2025can}, we encode the rich text information of user nodes into SentenceBert~\cite{reimers2019sentence} to obtain node attribute features. 
Detailed discussion about data construction is provided in Appendix~\ref{app: x}.

\subsubsection{Performance Comparison for Community Detection} 
Table~\ref{tab: X-hydrug-comm} reports the performance comparison among baseline methods on \textsc{X-HyDrug-Comm} dataset.
According to the results, we have the following observations.
(i) homogeneous GNNs, i.e., GCN and GAT, underperform HyGNNs on this dataset, which demonstrates the necessity of leveraging higher-order interactions among users for drug trafficking community detection.
(ii) Attention-based HyGNNs, i.e., HCHA and AllSet, have better performance in a higher training ratio (i.e., 50\% training data) compared to convolution-based HyGNNs, i.e., HGNN and HNHN, while showing comparable or worse performance in a lower training ratio (i.e., 10\% and 20\% training data).
This indicates that the attention mechanism may require more training data to learn effective representations.
(iii) ED-HNN consistently outperforms all baseline methods across all training ratios and evaluation metrics, which demonstrates the effectiveness of the equivariant and disentangled representation learning for drug trafficking community detection.

\subsubsection{Performance Comparison for Role Detection}
We conduct experiments on baseline methods over \textsc{X-HyDrug-Role} and report the results in Table~\ref{tab: X-hydrug-role}. 
We find the following results. 
(i) MLPs demonstrate competitive performance, outperforming GNNs in both evaluation metrics across all training ratios, which indicates that message passing on a naive clique-expansion graph may introduce redundant or noisy information for the role detection task.
(ii) HyGNNs consistently outperform GNNs, which indicates the importance of modeling higher-order interactions among users for role detection in drug trafficking networks.
(iii) ED-HNN consistently achieves the highest accuracy across all training ratios, reaching 68.55\%, 69.40\%, and 70.66\% at 10\%, 20\%, and 50\% training data, respectively.
This demonstrates that the equivariant and disentangled design of ED-HNN effectively captures the structural patterns relevant to role classification in drug trafficking networks.

For \textsc{X-MRDrug-Role}, we report the results in Table~\ref{tab: X-mrdrug-role}.
The following observations emerge from the results.
(i) Standard GNNs, including GCN and GAT, exhibit poor GMean performance, with GCN achieving only 33.84\% and GAT 26.87\% at a 10\% training ratio.
This indicates that naive message passing without explicit imbalance handling fails to capture minority class patterns, as the majority class dominates the aggregation process.
(ii) AD-GSMOTE consistently achieves the best performance across all settings and metrics, demonstrating that adaptive graph-based oversampling techniques are particularly effective for handling the severe class imbalance inherent in real-world drug trafficking role detection.

\subsection{Dietary: Opioid Misuse}

\textbf{Why this matters.}
Opioids are commonly used for pain relief, while opioids produce euphoria and therefore might be easily misused~\cite{quigley1996opioid, pasternak2014opiate}.
To tackle this, a common treatment is medication-assisted treatment (MAT)~\cite{connery2015medication, huang2026glen} that takes the opioid agonists such as methadone or buprenorphine in combination with counseling. 
While effective, the various induced side effects can often trigger opioid relapse and require timely
identification of at-risk individuals to intervene.
Emerging evidence has unveiled a compelling nexus between dietary behaviors and substance dependency~\cite{zhang2024mopi, kheradmand2020nutritional, jeynes2017importance,zhang2024diet,huang2026glen}, where individuals with opioid misuse often exhibit distinctive nutritional signatures, i.e., pronounced deficiencies in essential nutrients, compromised metabolic function, and aberrant eating behaviors.
These dietary fingerprints offer a promising avenue for proactive identification of at-risk individuals, particularly in latent cases where users remain unaware of their dependency or deliberately conceal their substance use.

\noindent\textbf{Our Contribution.}
To this end, we introduce the opioid misuse detection task, which aims to predict users who face opioid misuse through their dietary behaviors, such as food and ingredients taken, daily habits, etc.
Afterward, we construct \textsc{NHANES-Diet}, a heterogeneous graph that models the inherent relational structure of dietary data, i.e., the complex interactions among users, foods, ingredients, and dietary habits to effectively capture both individual consumption patterns and shared dietary behaviors across user populations, providing rich structural and semantic information.
    Crucially, \textsc{NHANES-Diet} enables the exploitation of advanced graph learning techniques to uncover subtle, complex correlations between dietary patterns and drug misuse, thereby unlocking novel predictive capabilities for early intervention.
    
    \subsubsection{Data Construction and Annotation.}
    We utilize NHANES data from 2003 to 2020 to construct \textsc{NHANES-Diet}.
As shown in Figure~\ref{fig: data}(c), it features five types of nodes, including user, food, habit,
ingredient, category, and four types of relationships such as \textit{user-eat-
food}, \textit{user-have-habit}, \textit{food-contain-ingredient}, and \textit{food-belong\_to-
category}. 
Opioid misuse users are identified by (i) records of illicit opioid drug use and (ii) records of prescription opioid medication use, which is a criterion commonly employed in the domain~\cite{gu2022prevalence}.
Detailed discussion about data construction and annotation is provided in Appendix~\ref{app: nhance}.

\subsubsection{Performance Comparison for Opioid Misuse Detection.}
Table~\ref{tab: nhance-diet-role} reports the performance comparison among baseline methods on \textsc{NHANES-Diet} dataset.
The following observations emerge from the results.
(i) HAN consistently achieves the best performance across all training ratios and evaluation metrics, with accuracy scores of 74.06\%, 75.10\%, and 75.90\% at 10\%, 20\%, and 50\% training data, respectively.
This demonstrates that hierarchical attention over heterogeneous graph structures effectively captures the complex relationships between dietary patterns and opioid misuse risk.
(ii) GAT serves as a strong runner-up at lower training ratios (10\% and 20\%), while R-GCN emerges as the second-best performer at 50\% training ratio.
This transition suggests that explicit relation-type modeling becomes increasingly beneficial as more labeled data becomes available, enabling the model to learn distinct aggregation patterns for different dietary relationships.
(iii) HetGNNs, including HAN and R-GCN, consistently outperform homogeneous GNNs such as GCN and GAT, highlighting the importance of modeling the diverse node types and edge types inherent in dietary data for opioid misuse detection.
(iv) HGMAE exhibits the lowest performance among all methods, with accuracy ranging from 63.17\% to 64.83\%, indicating that self-supervised pre-training objectives designed for general heterogeneous graphs may not align well with the downstream task of opioid misuse detection.

\section{Observations from \modelname}
In this section, we highlight the key insights from our analysis throughout the \modelname.
These observations are intended to guide both graph learning researchers in understanding method strengths and limitations, and public health practitioners in selecting appropriate approaches for opioid crisis applications.

\noindent\textbf{Obs. 1. Homogeneous Graph Projections Lose Critical Structural Information.}
A natural question when working with complex relational data is whether simple homogeneous graph representations suffice for opioid crisis applications.
Our experiments provide a decisive answer. Projecting heterogeneous or hypergraph structures onto homogeneous graphs causes substantial information loss, leading to {significantly degraded performance in detecting opioid-related risks.}
Specifically, GCN and GAT achieve 60.96\% and 61.76\% AUC, respectively, at a 10\% training ratio for the opioid overdose detection task, whereas HetGNNs, i.e., HAN and HGT, achieve 79.14\% and 79.81\%. 
Similarly, for the community detection task, GCN achieves only 39.09\% F1-Macro on a clique-expanded homogeneous projection, whereas ED-HNN achieves 57.62\%.
Moreover, MLPs also outperform GNNs in these settings, indicating that the issue lies not in the learning process over the graph structures but in the inappropriate structural representation for the opioid crisis domain.

This finding validates our core principle in designing \modelname, which is to provide datasets with rich heterogeneous and higher-order structures that reflect the real-world complexities of opioid crisis data, rather than relying on simplified homogeneous graph representations that obscure critical relational information.
For example, in healthcare settings, patient risk emerges from complex interactions among patients, prescribers, pharmacies, and medications, and collapsing these relationships into undifferentiated edges obscures the prescribing patterns indicative of potential overdose.
In social network settings, drug trafficking communities operate through group-wise collaborations that pairwise clique expansions fail to model comprehensively.
This observation highlights the importance of developing methods that explicitly model relation types and hyperedge structures rather than relying on homogeneous projections.
Moreover, the key takeaway is that investing in proper heterogeneous or hypergraph data pipelines is essential for effective risk detection and intervention.

\noindent\textbf{Obs. 2. Higher-Order Modeling Provides Consistent but Task-Dependent Gains.}
Our results demonstrate that HyGNNs consistently outperform GNNs across illicit drug trafficking detection tasks, with the magnitude of improvement varying by task, reflecting the nature of the underlying relational patterns across opioid crisis applications.
For the community detection task, ED-HNN achieves 57.62\% F1-Macro at a 10\% training ratio, a substantial improvement over GCN (39.09\%).
On \textsc{X-HyDrug-Role} for role detection, HyGNNs maintain their advantage, with ED-HNN reaching 68.55\% Accuracy compared to 64.87\% for GCN.

This finding reflects the nature of these opioid crisis intervention tasks. Community membership is defined by patterns of co-participation in trafficking activities, which hypergraphs naturally capture, whereas distinguishing individual roles, such as sellers versus buyers, depends more on behavioral signals in which structural advantages are less pronounced.
For practitioners combating online drug trafficking, hypergraph modeling should be adopted for both tasks; however, community disruption efforts will yield the greatest returns from higher-order structural modeling, while role identification task may also benefit from enriched user-level behavioral features.

\noindent\textbf{Obs. 3. Imbalance Handling Can Outweigh Architectural Sophistication.}
In opioid crisis applications, a key challenge is that target populations often remain minorities among the overall samples, leading to severe class imbalance that can undermine model performance and real-world utility.
Our experiments demonstrate that addressing this class imbalance at the data level can yield larger performance gains than adopting more sophisticated GNN architectures, with direct implications for the effectiveness of crisis response systems.
For the drug trafficking role detection task, at a 10\% training ratio, AD-GSMOTE achieves 61.68\% GMean and 70.58\% F1-Macro, while standard GNNs collapse to 33.84\% and 26.87\% GMean, respectively.
This approximately 28-35\% GMean deficit means that standard methods would miss the majority of actual drug sellers and buyers, severely limiting the utility of such systems for law enforcement and platform moderation.
We hope this finding motivates the graph learning community to prioritize developing graph learning methods that are robust to class imbalance while preserving relational semantics.
For practitioners deploying systems to combat the opioid crisis, where missing a single high-risk patient or trafficking network can have severe consequences, data-level interventions such as adaptive oversampling and class reweighting should be prioritized as essential components of any detection pipeline.

\section{Conclusion}
In this paper, we introduce \textsc{\modelname}, the first comprehensive benchmark for evaluating graph learning methods in the context of the opioid crisis.
\modelname provides five curated datasets spanning three critical application domains, namely opioid overdose detection from medical claims, illicit drug trafficking detection on digital platforms, and opioid misuse prediction from dietary patterns.
Through extensive experiments, we reveal actionable insights into the strengths and limitations of existing graph learning methods for opioid crisis applications.
Our benchmark contributes to the field in researchers and practitioners by providing a standardized evaluation framework and a rich set of datasets that mimic real-world scenarios, fostering the development of more effective graph-based methods to combat the opioid epidemic.
We release \modelname as an open-source library to support future research in combating the opioid epidemic.

\newpage
\bibliographystyle{ACM-Reference-Format}
\bibliography{citations}

@article{ye2025llms4all,
  title={Llms4all: A review of large language models across academic disciplines},
  author={Ye, Yanfang and Zhang, Zheyuan and Ma, Tianyi and Wang, Zehong and Li, Yiyang and Hou, Shifu and Sun, Weixiang and Shi, Kaiwen and Ma, Yijun and Song, Wei and others},
  journal={arXiv preprint arXiv:2509.19580},
  year={2025}
}

@article{yuan2024mora,
  title={Mora: Enabling generalist video generation via a multi-agent framework},
  author={Yuan, Zhengqing and Liu, Yixin and Cao, Yihan and Sun, Weixiang and Jia, Haolong and Chen, Ruoxi and Li, Zhaoxu and Lin, Bin and Yuan, Li and He, Lifang and others},
  journal={arXiv preprint arXiv:2403.13248},
  year={2024}
}

@inproceedings{fan2017social,
  title={Social media for opioid addiction epidemiology: Automatic detection of opioid addicts from twitter and case studies},
  author={Fan, Yujie and Zhang, Yiming and Ye, Yanfang and Li, Xin and Zheng, Wanhong},
  booktitle={Proceedings of the 2017 ACM on Conference on Information and Knowledge Management},
  pages={1259--1267},
  year={2017}
}

@article{ju2022grape,
  title={Grape: Knowledge graph enhanced passage reader for open-domain question answering},
  author={Ju, Mingxuan and Yu, Wenhao and Zhao, Tong and Zhang, Chuxu and Ye, Yanfang},
  journal={arXiv preprint arXiv:2210.02933},
  year={2022}
}

@inproceedings{zhao2023self,
  title={Self-supervised graph structure refinement for graph neural networks},
  author={Zhao, Jianan and Wen, Qianlong and Ju, Mingxuan and Zhang, Chuxu and Ye, Yanfang},
  booktitle={Proceedings of the sixteenth ACM international conference on web search and data mining},
  pages={159--167},
  year={2023}
}

@inproceedings{zhao2021multi,
  title={Multi-view self-supervised heterogeneous graph embedding},
  author={Zhao, Jianan and Wen, Qianlong and Sun, Shiyu and Ye, Yanfang and Zhang, Chuxu},
  booktitle={Joint European conference on machine learning and knowledge discovery in databases},
  pages={319--334},
  year={2021},
  organization={Springer}
}

@article{qian2022co,
  title={Co-modality graph contrastive learning for imbalanced node classification},
  author={Qian, Yiyue and Zhang, Chunhui and Zhang, Yiming and Wen, Qianlong and Ye, Yanfang and Zhang, Chuxu},
  journal={Advances in Neural Information Processing Systems},
  volume={35},
  pages={15862--15874},
  year={2022}
}

@inproceedings{ju2022adaptive,
  title={Adaptive kernel graph neural network},
  author={Ju, Mingxuan and Hou, Shifu and Fan, Yujie and Zhao, Jianan and Ye, Yanfang and Zhao, Liang},
  booktitle={Proceedings of the AAAI conference on artificial intelligence},
  volume={36},
  number={6},
  pages={7051--7058},
  year={2022}
}

@article{wang2024gft,
  title={Gft: Graph foundation model with transferable tree vocabulary},
  author={Wang, Zehong and Zhang, Zheyuan and Chawla, Nitesh and Zhang, Chuxu and Ye, Yanfang},
  journal={Advances in Neural Information Processing Systems},
  volume={37},
  pages={107403--107443},
  year={2024}
}

@artifactsoftware{R,
  title        = {R: A Language and Environment for Statistical Computing},
  author       = {{R Core Team}},
  organization = {R Foundation for Statistical Computing},
  address      = {Vienna, Austria},
  year         = {2019},
  url          = {https://www.R-project.org/}
}

@article{zhang2024mopi,
  title   = {MOPI-HFRS: A Multi-objective Personalized Health-aware Food Recommendation System with LLM-enhanced Interpretation},
  author  = {Zhang, Zheyuan and Wang, Zehong and Ma, Tianyi and Taneja, Varun Sameer and Nelson, Sofia and Le, Nhi Ha Lan and Murugesan, Keerthiram and Ju, Mingxuan and Chawla, Nitesh V and Zhang, Chuxu and others},
  journal = {arXiv preprint arXiv:2412.08847},
  year    = {2024}
}

@article{reimers2019sentence,
  title   = {Sentence-bert: Sentence embeddings using siamese bert-networks},
  author  = {Reimers, Nils and Gurevych, Iryna},
  journal = {arXiv preprint arXiv:1908.10084},
  year    = {2019}
}

@inproceedings{feng2019hypergraph,
  title     = {Hypergraph neural networks},
  author    = {Feng, Yifan and You, Haoxuan and Zhang, Zizhao and Ji, Rongrong and Gao, Yue},
  booktitle = {AAAI},
  year      = {2019}
}

@inproceedings{chien2021you,
  title     = {You are allset: A multiset function framework for hypergraph neural networks},
  author    = {Chien, Eli and Pan, Chao and Peng, Jianhao and Milenkovic, Olgica},
  booktitle = {ICLR},
  year      = {2022}
}

@inproceedings{wang2022equivariant,
  title     = {Equivariant hypergraph diffusion neural operators},
  author    = {Wang, Peihao and Yang, Shenghao and Liu, Yunyu and Wang, Zhangyang and Li, Pan},
  booktitle = {ICLR},
  year      = {2023}
}

@misc{cdc_understanding_opioid_overdose_epidemic,
  title        = {Understanding the Opioid Overdose Epidemic},
  author       = {{Centers for Disease Control and Prevention}},
  howpublished = {Centers for Disease Control and Prevention},
  note         = {National Center for Injury Prevention and Control},
  year         = {2025},
  month        = {June},
  url          = {https://www.cdc.gov/overdose-prevention/about/understanding-the-opioid-overdose-epidemic.html},
  urldate      = {2025-12-18}
}

@article{dong2023integrated,
  title     = {An integrated LSTM-HeteroRGNN model for interpretable opioid overdose risk prediction},
  author    = {Dong, Xinyu and Wong, Rachel and Lyu, Weimin and Abell-Hart, Kayley and Deng, Jianyuan and Liu, Yinan and Hajagos, Janos G and Rosenthal, Richard N and Chen, Chao and Wang, Fusheng},
  journal   = {Artificial intelligence in medicine},
  volume    = {135},
  pages     = {102439},
  year      = {2023},
  publisher = {Elsevier}
}

@inproceedings{wen2022disentangled,
  title     = {Disentangled dynamic heterogeneous graph learning for opioid overdose prediction},
  author    = {Wen, Qianlong and Ouyang, Zhongyu and Zhang, Jianfei and Qian, Yiyue and Ye, Yanfang and Zhang, Chuxu},
  booktitle = {KDD},
  year      = {2022}
}

@inproceedings{rosman2023detecting,
  title     = {Detecting anomalous networks of opioid prescribers and dispensers in prescription drug data},
  author    = {Rosman, Katie and Neill, Daniel B},
  booktitle = {AAAI},
  year      = {2023}
}

@inproceedings{ma2023hypergraph,
  title     = {Hypergraph contrastive learning for drug trafficking community detection},
  author    = {Ma, Tianyi and Qian, Yiyue and Zhang, Chuxu and Ye, Yanfang},
  booktitle = {ICDM},
  year      = {2023}
}

@inproceedings{ma2025llm,
  title     = {Llm-empowered class imbalanced graph prompt learning for online drug trafficking detection},
  author    = {Ma, Tianyi and Qian, Yiyue and Wang, Zehong and Zhang, Zheyuan and Zhang, Chuxu and Ye, Yanfang},
  booktitle = {Findings of ACL},
  year      = {2025}
}

@inproceedings{kipf2017semisupervisedclassificationgraphconvolutional,
  title     = {Semi-Supervised Classification with Graph Convolutional Networks},
  author    = {Thomas N. Kipf and Max Welling},
  year      = {2017},
  booktitle = {ICLR}
}

@inproceedings{velivckovic2017graph,
  title     = {Graph attention networks},
  author    = {Veli{\v{c}}kovi{\'c}, Petar and Cucurull, Guillem and Casanova, Arantxa and Romero, Adriana and Lio, Pietro and Bengio, Yoshua},
  booktitle = {ICLR},
  year      = {2018}
}

@inproceedings{hamilton2017inductive,
  title     = {Inductive representation learning on large graphs},
  author    = {Hamilton, Will and Ying, Zhitao and Leskovec, Jure},
  booktitle = {NeurIPS},
  year      = {2017}
}

@inproceedings{qian2021distilling,
  title     = {Distilling meta knowledge on heterogeneous graph for illicit drug trafficker detection on social media},
  author    = {Qian, Yiyue and Zhang, Yiming and Ye, Yanfang and Zhang, Chuxu and others},
  booktitle = {NeurIPS},
  year      = {2021}
}

@article{zheng2015oversampling,
  title   = {Oversampling method for imbalanced classification},
  author  = {Zheng, Zhuoyuan and Cai, Yunpeng and Li, Ye},
  journal = {Computing and Informatics},
  volume  = {34},
  number  = {5},
  pages   = {1017--1037},
  year    = {2015}
}

@article{chawla2002smote,
  title   = {SMOTE: synthetic minority over-sampling technique},
  author  = {Chawla, Nitesh V and Bowyer, Kevin W and Hall, Lawrence O and Kegelmeyer, W Philip},
  journal = {Journal of artificial intelligence research},
  volume  = {16},
  pages   = {321--357},
  year    = {2002}
}

@inproceedings{schlichtkrull2018modeling,
  title     = {Modeling relational data with graph convolutional networks},
  author    = {Schlichtkrull, Michael and Kipf, Thomas N and Bloem, Peter and Van Den Berg, Rianne and Titov, Ivan and Welling, Max},
  booktitle = {ESWC},
  year      = {2018}
}

@inproceedings{tian2023heterogeneous,
  title     = {Heterogeneous graph masked autoencoders},
  author    = {Tian, Yijun and Dong, Kaiwen and Zhang, Chunhui and Zhang, Chuxu and Chawla, Nitesh V},
  booktitle = {AAAI},
  year      = {2023}
}

@inproceedings{hu2020heterogeneous,
  title     = {Heterogeneous graph transformer},
  author    = {Hu, Ziniu and Dong, Yuxiao and Wang, Kuansan and Sun, Yizhou},
  booktitle = {WWW},
  year      = {2020}
}

@inproceedings{wang2019heterogeneous,
  title     = {Heterogeneous graph attention network},
  author    = {Wang, Xiao and Ji, Houye and Shi, Chuan and Wang, Bai and Ye, Yanfang and Cui, Peng and Yu, Philip S},
  booktitle = {WWW},
  year      = {2019}
}

@article{dong2020hnhn,
  title   = {Hnhn: Hypergraph networks with hyperedge neurons},
  author  = {Dong, Yihe and Sawin, Will and Bengio, Yoshua},
  journal = {arXiv preprint arXiv:2006.12278},
  year    = {2020}
}

@article{bai2021hypergraph,
  title     = {Hypergraph convolution and hypergraph attention},
  author    = {Bai, Song and Zhang, Feihu and Torr, Philip HS},
  journal   = {Pattern Recognition},
  volume    = {110},
  pages     = {107637},
  year      = {2021},
  publisher = {Elsevier}
}

@inproceedings{park2021graphens,
  title     = {Graphens: Neighbor-aware ego network synthesis for class-imbalanced node classification},
  author    = {Park, Joonhyung and Song, Jaeyun and Yang, Eunho},
  booktitle = {ICLR},
  year      = {2021}
}

@inproceedings{zhao2021graphsmote,
  title     = {Graphsmote: Imbalanced node classification on graphs with graph neural networks},
  author    = {Zhao, Tianxiang and Zhang, Xiang and Wang, Suhang},
  booktitle = {WSDM},
  year      = {2021}
}

@inproceedings{qian2025adaptive,
  title     = {Adaptive graph enhancement for imbalanced multi-relation graph learning},
  author    = {Qian, Yiyue and Ma, Tianyi and Zhang, Chuxu and Ye, Yanfang},
  booktitle = {WSDM},
  year      = {2025}
}

@article{lossio2022opioid2mme,
  title     = {Opioid2MME: standardizing opioid prescriptions to morphine milligram equivalents from electronic health records},
  author    = {Lossio-Ventura, Juan Antonio and Song, Wenyu and Sainlaire, Michael and Dykes, Patricia C and Hernandez-Boussard, Tina},
  journal   = {International journal of medical informatics},
  volume    = {162},
  pages     = {104739},
  year      = {2022},
  publisher = {Elsevier}
}

@article{adams2025standardizing,
  title     = {Standardizing research methods for opioid dose comparison: the NIH HEAL morphine milligram equivalent calculator},
  author    = {Adams, Meredith CB and Sward, Katherine A and Perkins, Matthew L and Hurley, Robert W},
  journal   = {Pain},
  volume    = {166},
  number    = {8},
  pages     = {1729--1737},
  year      = {2025},
  publisher = {LWW}
}

@article{dowell2022cdc,
  title   = {CDC clinical practice guideline for prescribing opioids for pain—United States, 2022},
  author  = {Dowell, Deborah},
  journal = {MMWR. Recommendations and reports},
  volume  = {71},
  year    = {2022}
}

@article{gu2022prevalence,
  title     = {Prevalence of self-reported prescription opioid use and illicit drug use among US adults: NHANES 2005--2016},
  author    = {Gu, Ja K and Allison, Penelope and Trotter, Alexis Grimes and Charles, Luenda E and Ma, Claudia C and Groenewold, Matthew and Andrew, Michael E and Luckhaupt, Sara E},
  journal   = {Journal of occupational and environmental medicine},
  volume    = {64},
  number    = {1},
  pages     = {39--45},
  year      = {2022},
  publisher = {LWW}
}

@misc{who2023urgent,
  author = {{World Health Organization}},
  title  = {Urgent action needed to tackle stalled progress on health-related Sustainable Development Goals},
  year   = {2023},
  url    = {https://www.who.int/news/item/19-05-2023-urgent-action-needed-to-tackle-stalled-progress-on-health-related-sustainable-development-goals}
}

@article{florence2021economic,
  title     = {The economic burden of opioid use disorder and fatal opioid overdose in the United States, 2017},
  author    = {Florence, Curtis and Luo, Feijun and Rice, Ketra},
  journal   = {Drug and alcohol dependence},
  volume    = {218},
  pages     = {108350},
  year      = {2021},
  publisher = {Elsevier}
}

@article{maclean2020economic,
  title     = {Economic studies on the opioid crisis: A review},
  author    = {Maclean, Johanna Catherine and Mallatt, Justine and Ruhm, Christopher J and Simon, Kosali},
  year      = {2020},
  publisher = {National Bureau of Economic Research}
}

@article{brundage2019ripple,
  title   = {The ripple effect: The impact of the opioid epidemic on children and families},
  author  = {Brundage, Suzanne C and Levine, Carol},
  journal = {New York, NY: United Hospital Fund},
  volume  = {46},
  year    = {2019}
}

@article{kamdar2019knowledge,
  title   = {A knowledge graph-based approach for exploring the US opioid epidemic},
  author  = {Kamdar, Maulik R and Hamamsy, Tymor and Shelton, Shea and Vala, Ayin and Eftimov, Tome and Zou, James and Tamang, Suzanne},
  journal = {arXiv preprint arXiv:1905.11513},
  year    = {2019}
}

@inproceedings{fan2018automatic,
  title     = {Automatic Opioid User Detection from Twitter: Transductive Ensemble Built on Different Meta-graph Based Similarities over Heterogeneous Information Network.},
  author    = {Fan, Yujie and Zhang, Yiming and Ye, Yanfang and Li, Xin},
  booktitle = {IJCAI},
  year      = {2018}
}

@article{rana2025beyond,
  title     = {Beyond REMS \& PDMPs: A Proposed Framework for Next-Generation Opioid Regulation},
  author    = {Rana, Aysha and Ram, Kavetha},
  journal   = {Therapeutic Innovation \& Regulatory Science},
  pages     = {1--12},
  year      = {2025},
  publisher = {Springer}
}

@article{singh2019should,
  title     = {How should medical education better prepare physicians for opioid prescribing?},
  author    = {Singh, Rohanit and Pushkin, Gary W},
  journal   = {AMA Journal of Ethics},
  volume    = {21},
  number    = {8},
  pages     = {636--641},
  year      = {2019},
  publisher = {American Medical Association}
}

@article{lavorgna2014internet,
  title     = {Internet-mediated drug trafficking: towards a better understanding of new criminal dynamics},
  author    = {Lavorgna, Anita},
  journal   = {Trends in organized crime},
  volume    = {17},
  number    = {4},
  pages     = {250--270},
  year      = {2014},
  publisher = {Springer}
}

@article{lavorgna2016use,
  title     = {How the use of the internet is affecting drug trafficking practices},
  author    = {Lavorgna, Anita},
  year      = {2016},
  publisher = {European Monitoring Centre for Drugs and Drug Addiction}
}

@article{broseus2016studying,
  title     = {Studying illicit drug trafficking on Darknet markets: structure and organisation from a Canadian perspective},
  author    = {Bros{\'e}us, Julian and Rhumorbarbe, Damien and Mireault, Caroline and Ouellette, Vincent and Crispino, Frank and D{\'e}cary-H{\'e}tu, David},
  journal   = {Forensic science international},
  volume    = {264},
  pages     = {7--14},
  year      = {2016},
  publisher = {Elsevier}
}

@article{chavez2020nutritional,
  title     = {Nutritional implications of opioid use disorder: A guide for drug treatment providers.},
  author    = {Chavez, Melody N and Rigg, Khary K},
  journal   = {Psychology of Addictive Behaviors},
  volume    = {34},
  number    = {6},
  pages     = {699},
  year      = {2020},
  publisher = {American Psychological Association}
}

@article{matos2020opioids,
  title     = {Opioids, polypharmacy, and drug interactions: A technological paradigm shift is needed to ameliorate the ongoing opioid epidemic},
  author    = {Matos, Adriana and Bankes, David L and Bain, Kevin T and Ballinghoff, Tyler and Turgeon, Jacques},
  journal   = {Pharmacy},
  volume    = {8},
  number    = {3},
  pages     = {154},
  year      = {2020},
  publisher = {MDPI}
}

@article{hughes2017social,
  title     = {Social network analysis of Australian poly-drug trafficking networks: How do drug traffickers manage multiple illicit drugs?},
  author    = {Hughes, Caitlin E and Bright, David A and Chalmers, Jenny},
  journal   = {Social Networks},
  volume    = {51},
  pages     = {135--147},
  year      = {2017},
  publisher = {Elsevier}
}

@inproceedings{goel2023llms,
  title     = {Llms accelerate annotation for medical information extraction},
  author    = {Goel, Akshay and Gueta, Almog and Gilon, Omry and Liu, Chang and Erell, Sofia and Nguyen, Lan Huong and Hao, Xiaohong and Jaber, Bolous and Reddy, Shashir and Kartha, Rupesh and others},
  booktitle = {ML4H},
  year      = {2023}
}

@inproceedings{wei2018clinical,
  title     = {Clinical text annotation--what factors are associated with the cost of time?},
  author    = {Wei, Qiang and Franklin, Amy and Cohen, Trevor and Xu, Hua},
  booktitle = {AMIA},
  volume    = {2018},
  pages     = {1552},
  year      = {2018}
}

@inproceedings{hu2021detection,
  title     = {Detection of illicit drug trafficking events on instagram: A deep multimodal multilabel learning approach},
  author    = {Hu, Chuanbo and Yin, Minglei and Liu, Bin and Li, Xin and Ye, Yanfang},
  booktitle = {CIKM},
  year      = {2021}
}

@article{hu2021identifying,
  title   = {Identifying illicit drug dealers on instagram with large-scale multimodal data fusion},
  author  = {Hu, Chuanbo and Yin, Minglei and Liu, Bin and Li, Xin and Ye, Yanfang},
  journal = {ACM Transactions on Intelligent Systems and Technology (TIST)},
  year    = {2021}
}

@inproceedings{hu2020open,
  title     = {Open graph benchmark: Datasets for machine learning on graphs},
  author    = {Hu, Weihua and Fey, Matthias and Zitnik, Marinka and Dong, Yuxiao and Ren, Hongyu and Liu, Bowen and Catasta, Michele and Leskovec, Jure},
  booktitle = {NeurIPS},
  year      = {2020}
}

@inproceedings{dwivedi2022long,
  title     = {Long range graph benchmark},
  author    = {Dwivedi, Vijay Prakash and Ramp{\'a}{\v{s}}ek, Ladislav and Galkin, Michael and Parviz, Ali and Wolf, Guy and Luu, Anh Tuan and Beaini, Dominique},
  booktitle = {NeurIPS},
  year      = {2022}
}

@inproceedings{li2023gslb,
  title     = {GSLB: the graph structure learning benchmark},
  author    = {Li, Zhixun and Wang, Liang and Sun, Xin and Luo, Yifan and Zhu, Yanqiao and Chen, Dingshuo and Luo, Yingtao and Zhou, Xiangxin and Liu, Qiang and Wu, Shu and others},
  booktitle = {NeurIPS},
  year      = {2023}
}

@misc{nhance,
  title  = {National Health and Nutrition Examination Survey},
  year   = 2024,
  author = {CDC},
  url    = {https://wwwn.cdc.gov/nchs/nhanes/default.aspx}
}

@article{mcquay1999opioids,
  title     = {Opioids in pain management},
  author    = {McQuay, Henry},
  journal   = {The Lancet},
  volume    = {353},
  number    = {9171},
  pages     = {2229--2232},
  year      = {1999},
  publisher = {Elsevier}
}

@article{mehta2006acute,
  title     = {Acute pain management for opioid dependent patients},
  author    = {Mehta, V and Langford, RM},
  journal   = {Anaesthesia},
  volume    = {61},
  number    = {3},
  pages     = {269--276},
  year      = {2006},
  publisher = {Wiley Online Library}
}

@article{judd2023opioid,
  title     = {The opioid epidemic: a review of the contributing factors, negative consequences, and best practices},
  author    = {Judd, Dallin and King, Connor R and Galke, Curtis and King, Connor and Galke, Curtis},
  journal   = {Cureus},
  volume    = {15},
  number    = {7},
  year      = {2023},
  publisher = {Cureus}
}

@article{chen2019prevention,
  title     = {Prevention of prescription opioid misuse and projected overdose deaths in the United States},
  author    = {Chen, Qiushi and Larochelle, Marc R and Weaver, Davis T and Lietz, Anna P and Mueller, Peter P and Mercaldo, Sarah and Wakeman, Sarah E and Freedberg, Kenneth A and Raphel, Tiana J and Knudsen, Amy B and others},
  journal   = {JAMA network open},
  volume    = {2},
  number    = {2},
  pages     = {e187621--e187621},
  year      = {2019},
  publisher = {American Medical Association}
}

@misc{carrillo2022lancet,
  title  = {Lancet Regional Health Americas},
  author = {Carrillo-Larco, Rodrigo M and Guzm{\'a}n-Vilca, Wilmer Cristobal and Le{\'o}n-Velarde, Fabiola and Bernabe-Ortiz, Antonio and Jimenez, M Michelle and Penny, Mary E and Gianella, Camila and Legu{\'\i}a, Mariana and Tsukayama, Pablo and Hartinger, Stella M and others},
  year   = {2022}
}

@article{quigley1996opioid,
  title     = {Opioid switching to improve pain relief and drug tolerability},
  author    = {Quigley, Columba and Cochrane Pain, Palliative and Supportive Care Group},
  journal   = {Cochrane Database of Systematic Reviews},
  volume    = {2010},
  number    = {11},
  year      = {1996},
  publisher = {John Wiley \& Sons, Ltd Chichester, UK}
}

@article{pasternak2014opiate,
  title     = {Opiate pharmacology and relief of pain},
  author    = {Pasternak, Gavril W},
  journal   = {Journal of Clinical Oncology},
  volume    = {32},
  number    = {16},
  pages     = {1655--1661},
  year      = {2014},
  publisher = {American Society of Clinical Oncology}
}

@article{connery2015medication,
  title     = {Medication-assisted treatment of opioid use disorder: review of the evidence and future directions},
  author    = {Connery, Hilary Smith},
  journal   = {Harvard review of psychiatry},
  volume    = {23},
  number    = {2},
  pages     = {63--75},
  year      = {2015},
  publisher = {LWW}
}

@article{kheradmand2020nutritional,
  title     = {Nutritional status in patients under methadone maintenance treatment},
  author    = {Kheradmand, Ali and Kheradmand, Azadeh},
  journal   = {Journal of Substance Use},
  volume    = {25},
  number    = {2},
  pages     = {173--176},
  year      = {2020},
  publisher = {Taylor \& Francis}
}

@article{jeynes2017importance,
  title     = {The importance of nutrition in aiding recovery from substance use disorders: A review},
  author    = {Jeynes, Kendall D and Gibson, E Leigh},
  journal   = {Drug and alcohol dependence},
  volume    = {179},
  pages     = {229--239},
  year      = {2017},
  publisher = {Elsevier}
}

@article{garcia2009labor,
  title   = {Labor migration, drug trafficking organizations, and drug use: major challenges for transnational communities in Mexico},
  author  = {Garc{\'\i}a, Victor and Gonz{\'a}lez, Laura},
  journal = {Urban anthropology and studies of cultural systems and world economic development},
  volume  = {38},
  number  = {2-4},
  pages   = {303},
  year    = {2009}
}

@article{browne2003drug,
  title     = {Drug supply and trafficking: An overview},
  author    = {Browne, Deborah and Mason, Mark and Murphy, Rachel},
  journal   = {How. J. Crim. Just.},
  volume    = {42},
  pages     = {324},
  year      = {2003},
  publisher = {HeinOnline}
}

@article{imlearn,
  author  = {Guillaume  Lema{{\^i}}tre and Fernando Nogueira and Christos K. Aridas},
  title   = {Imbalanced-learn: A Python Toolbox to Tackle the Curse of Imbalanced Datasets in Machine Learning},
  journal = {Journal of Machine Learning Research},
  year    = {2017},
  volume  = {18},
  number  = {17},
  pages   = {1-5},
  url     = {http://jmlr.org/papers/v18/16-365}
}

@inproceedings{feyetal2025,
  title     = {PyG 2.0: Scalable Learning on Real World Graphs},
  author    = {Fey, Matthias and Sunil, Jinu and Nitta, Akihiro and Puri, Rishi and Shah, Manan and Stojanovic, Blaz and Bendias, Ramona and Alexandria, Barghi and Kocijan, Vid and Zhang, Zecheng and He, Xinwei and Lenssen, Jan E. and Leskovec, Jure},
  booktitle = {Temporal Graph Learning Workshop at KDD},
  year      = {2025}
}

@article{dwyer2003estimation,
  title={Estimation of usual intakes: what we eat in America--NHANES},
  author={Dwyer, Johanna and Picciano, Mary Frances and Raiten, Daniel J and Steering Committee and others},
  journal={The Journal of nutrition},
  volume={133},
  number={2},
  pages={609S--623S},
  year={2003},
  publisher={Elsevier}
}

@article{montville2013usda,
  title={USDA food and nutrient database for dietary studies (FNDDS), 5.0},
  author={Montville, Janice B and Ahuja, Jaspreet KC and Martin, Carrie L and Heendeniya, Kaushalya Y and Omolewa-Tomobi, Grace and Steinfeldt, Lois C and Anand, Jaswinder and Adler, Meghan E and LaComb, Randy P and Moshfegh, Alanna},
  journal={Procedia Food Science},
  volume={2},
  pages={99--112},
  year={2013},
  publisher={Elsevier}
}

@inproceedings{zhang2019heterogeneous,
  title={Heterogeneous graph neural network},
  author={Zhang, Chuxu and Song, Dongjin and Huang, Chao and Swami, Ananthram and Chawla, Nitesh V},
  booktitle={KDD},
  year={2019}
}

@article{huang2026glen,
  title={GLEN-Bench: A Graph-Language based Benchmark for Nutritional Health},
  author={Huang, Jiatan and Zhang, Zheyuan and Ma, Tianyi and Li, Mingchen and Zheng, Yaning and Ye, Yanfang and Zhang, Chuxu},
  journal={arXiv preprint arXiv:2601.18106},
  year={2026}
}

@inproceedings{zhang2024diet,
  title={Diet-odin: A novel framework for opioid misuse detection with interpretable dietary patterns},
  author={Zhang, Zheyuan and Wang, Zehong and Hou, Shifu and Hall, Evan and Bachman, Landon and White, Jasmine and Galassi, Vincent and Chawla, Nitesh V and Zhang, Chuxu and Ye, Yanfang},
  booktitle={KDD},
  year={2024}
}

@article{wang2025beyond,
  title={Beyond message passing: Neural graph pattern machine},
  author={Wang, Zehong and Zhang, Zheyuan and Ma, Tianyi and Chawla, Nitesh V and Zhang, Chuxu and Ye, Yanfang},
  journal={arXiv preprint arXiv:2501.18739},
  year={2025}
}

@article{wang2025graph,
  title={Graph foundation models: A comprehensive survey},
  author={Wang, Zehong and Liu, Zheyuan and Ma, Tianyi and Li, Jiazheng and Zhang, Zheyuan and Fu, Xingbo and Li, Yiyang and Yuan, Zhengqing and Song, Wei and Ma, Yijun and others},
  journal={arXiv preprint arXiv:2505.15116},
  year={2025}
}

@inproceedings{wang2025can,
  title={Can LLMs convert graphs to text-attributed graphs?},
  author={Wang, Zehong and Liu, Sidney and Zhang, Zheyuan and Ma, Tianyi and Zhang, Chuxu and Ye, Yanfang},
  booktitle={NAACL},
  year={2025}
}

@inproceedings{wang2024towards,
  title={Towards graph foundation models: Learning generalities across graphs via task-trees},
  author={Wang, Zehong and Zhang, Zheyuan and Ma, Tianyi and Chawla, Nitesh V and Zhang, Chuxu and Ye, Yanfang},
  booktitle={ICML},
  year={2025}
}

@inproceedings{zhang2025ngqa,
  title={Ngqa: a nutritional graph question answering benchmark for personalized health-aware nutritional reasoning},
  author={Zhang, Zheyuan and Li, Yiyang and Le, Nhi Ha Lan and Wang, Zehong and Ma, Tianyi and Galassi, Vincent and Murugesan, Keerthiram and Moniz, Nuno and Geyer, Werner and Chawla, Nitesh V and others},
  booktitle={ACL},
  year={2025}
}

@article{ma2025adaptive,
  title={Adaptive expansion for hypergraph learning},
  author={Ma, Tianyi and Qian, Yiyue and Zhang, Shinan and Zhang, Chuxu and Ye, Yanfang},
  journal={arXiv preprint arXiv:2502.15564},
  year={2025}
}

@article{zhang2025agentrouter,
  title={AgentRouter: A Knowledge-Graph-Guided LLM Router for Collaborative Multi-Agent Question Answering},
  author={Zhang, Zheyuan and Shi, Kaiwen and Yuan, Zhengqing and Wang, Zehong and Ma, Tianyi and Murugesan, Keerthiram and Galassi, Vincent and Zhang, Chuxu and Ye, Yanfang},
  journal={arXiv preprint arXiv:2510.05445},
  year={2025}
}

@inproceedings{wang2025generative,
  title={Generative graph pattern machine},
  author={Wang, Zehong and Zhang, Zheyuan and Ma, Tianyi and Zhang, Chuxu and Ye, Yanfang},
  booktitle={NeurIPS},
  year={2025}
}

@article{ma2025psyscam,
  title={PsyScam: a benchmark for psychological techniques in real-world scams},
  author={Ma, Shang and Ma, Tianyi and Liu, Jiahao and Song, Wei and Liang, Zhenkai and Xiao, Xusheng and Ye, Yanfang},
  journal={arXiv preprint arXiv:2505.15017},
  year={2025}
}

@inproceedings{ma2025autodata,
  title={Autodata: A multi-agent system for open web data collection},
  author={Ma, Tianyi and Qian, Yiyue and Zhang, Zheyuan and Wang, Zehong and Qian, Xiaoye and Bai, Feifan and Ding, Yifan and Luo, Xuwei and Zhang, Shinan and Murugesan, Keerthiram and others},
  booktitle={NeurIPS},
  year={2025}
}

@article{wang2026molecular,
  title={Molecular Representations in Implicit Functional Space via Hyper-Networks},
  author={Wang, Zehong and Han, Xiaolong and Yang, Qi and Tang, Xiangru and Wu, Fang and Guo, Xiaoguang and Sun, Weixiang and Ma, Tianyi and Lio, Pietro and Cong, Le and others},
  journal={arXiv preprint arXiv:2601.22327},
  year={2026}
}

@article{li2026longda,
  title={LongDA: Benchmarking LLM Agents for Long-Document Data Analysis},
  author={Li, Yiyang and Zhang, Zheyuan and Ma, Tianyi and Wang, Zehong and Murugesan, Keerthiram and Zhang, Chuxu and Ye, Yanfang},
  journal={arXiv preprint arXiv:2601.02598},
  year={2026}
}

@inproceedings{ma202586,
  title={Hypergraph Representation Learning with Adaptive Broadcasting and Receiving},
  author={Ma, Tianyi and Qian, Yiyue and Wang, Zehong and Zhang, Zheyuan and Zhang, Shinan and Zhang, Chuxu and Ye, Fanny},
  booktitle={ICDM},
  year={2025}
}

\appendix 

\section{Baseline Implementation Details} \label{app: baseline}
Our benchmark framework supports all graph learning methods seamlessly.
We implement classic baseline methods in the literature, including feature-based methods (G1), graph neural networks (G2), heterogeneous graph neural networks (G3), multi-relational graph neural networks (G4), and hypergraph neural networks (G5).
In this section, we provide detailed descriptions and implementation details for each method.

\noindent\textbf{G1. Feature-based Methods.} 
We implement the following feature-based methods as baselines that ignore the graph structure, serving as important references to evaluate whether graph-based methods provide meaningful improvements over structure-agnostic approaches.

\begin{itemize}[leftmargin=*]
    \item \textbf{MLP} serves as the foundational baseline that processes node features independently without considering graph topology. 
    For each layer, it applies a series of linear transformations, followed by nonlinear activations, to produce node representations.
    \item \textbf{Oversampling}~\cite{zheng2015oversampling} is a feature-based method that addresses class imbalance by randomly creating samples from minority classes to match the majority class distribution.
    \item \textbf{SMOTE}~\cite{chawla2002smote} generates synthetic samples for minority classes by interpolating between existing minority samples and their k-nearest neighbors in the feature space.
\end{itemize}
For Oversampling and SMOTE, we apply these techniques on the training set while preserving the original class distribution in validation and test sets to ensure unbiased evaluation.
Moreover, we obtain source code for these methods from the widely used imbalanced-learn library~\cite{imlearn} to ensure standardized, reproducible code.

\noindent\textbf{G2. Graph Neural Networks (GNNs).} 
We implement the following GNNs as baselines for homogeneous graph learning. 
\begin{itemize}[leftmargin=*]
    \item \textbf{GCN}~\cite{kipf2017semisupervisedclassificationgraphconvolutional} performs spectral graph convolutions through a first-order approximation of localized spectral filters.
    GCN aggregates information from neighboring nodes through symmetric normalization, making it a fundamental baseline for evaluating graph structure utilization.
    
    \item \textbf{GAT}~\cite{velivckovic2017graph} introduces attention mechanisms with coefficient computation to learn adaptive importance weights for different neighbors during message passing.
    \item \textbf{GraphSAGE}~\cite{hamilton2017inductive} learns node embeddings through sampling and aggregating features from a fixed-size neighborhood, enabling inductive learning on unseen nodes.
    We use the mean aggregator in our implementation for computational efficiency and stable performance.
    \item \textbf{GraphSMOTE}~\cite{zhao2021graphsmote} extends SMOTE to graph-structured data by generating synthetic nodes for minority classes while preserving graph topology.
    It first embeds nodes using a GNN encoder, then generates synthetic minority nodes in the embedding space through interpolation, and finally reconstructs edges for synthetic nodes based on learned edge predictors.
\end{itemize}
To run GNNs over heterogeneous graphs, we ignore node and edge types and treat all nodes and edges as homogeneous, allowing us to evaluate the effectiveness of GNNs without explicit modeling of heterogeneity.
Following existing work~\cite {chien2021you, ma2025adaptive}, for the hypergraph datasets \textsc{X-HyDrug-Comm} and \textsc{X-HyDrug-Role}, we apply clique expansion to convert each hyperedge into pairwise edges between connected node pairs.
We implement these methods based on the PyG library~\cite{feyetal2025}.

\noindent\textbf{G3. Heterogeneous Graph Neural Networks (HetGNNs).} 
We implement the following HetGNNs as baselines for heterogeneous graph learning, which explicitly model multiple node and edge types to capture the rich semantic information in opioid crisis data.
\begin{itemize}[leftmargin=*]
    \item \textbf{HAN}~\cite{wang2019heterogeneous}  employs a hierarchical attention mechanism consisting of node-level and semantic-level attention.
    Node-level attention learns the importance ofmeta-path-basedd neighbors, while semantic-level attention learns the importance of different meta-paths.
    \item \textbf{HGT}~\cite{hu2020heterogeneous} adapts the Transformer architecture to heterogeneous graphs by introducing type-specific linear projections and relation-aware attention.
    \item \textbf{HGMAE}~\cite{tian2023heterogeneous} is a self-supervised pre-training method that learns node representations by reconstructing masked node features and graph structures.
    HGMAE first masks a portion of nodes and their associated edges, then uses a heterogeneous graph encoder to embed the corrupted graph, and finally reconstructs the masked content through type-specific decoders.
    The pre-trained encoder is then fine-tuned for downstream tasks.
\end{itemize}
We implement these methods using the source code from their original papers to ensure rigorous, reproducible execution.
For metapath-based methods, we design metapaths that capture meaningful relationships in the networks, and list them in Table~\ref{tab:pdmp_stat} and Table~\ref{tab:nhance_stat} for datasets \textsc{Pdmp-OD-Det} and \textsc{NHANES-Diet}, respectively.

\noindent\textbf{G4. Multi-Relational Graph Neural Networks (MRGNNs).} 
We implement the following MRGNNs as baselines for multi-relational graph learning, which handle graphs with a single node type but multiple edge types representing different interaction semantics.
\begin{itemize}[leftmargin=*]
    \item \textbf{R-GCN}~\cite{schlichtkrull2018modeling} extends GCN to multi-relational graphs by learning relation-specific transformation matrices.
    \item \textbf{GraphENS}~\cite{park2021graphens} addresses class imbalance through a neighbor sampling strategy that considers both node features and local topology.
    GraphENS synthesizes ego networks for minority-class nodes by interpolating between existing minority samples and their structural neighborhoods, thereby preserving local graph structure during augmentation.
    \item \textbf{AD-GSMOTE}~\cite{qian2025adaptive} extends GraphSMOTE with an adaptive mechanism that dynamically adjusts the synthesis strategy based on local graph density and class distribution.
\end{itemize}

\noindent\textbf{G5. Hypergraph Neural Networks (HyGNNs).} 
We implement the following HyGNNs as baselines for hypergraph learning, which capture higher-order interactions among multiple nodes through hyperedges. 

\begin{itemize}[leftmargin=*]
    \item \textbf{HGNN}~\cite{feng2019hypergraph} generalizes spectral graph convolution to hypergraphs using the hypergraph Laplacian.
    
    \item \textbf{HNHN}~\cite{dong2020hnhn} introduces learnable hyperedge representations that evolve alongside node representations.
    By maintaining explicit hyperedge embeddings, HNHN captures the semantics of group interactions within drug-trafficking communities.
    
    \item \textbf{HCHA}~\cite{bai2021hypergraph} combines spectral hypergraph convolution with attention mechanisms to learn adaptive hyperedge importance, enabling the model to focus on the most relevant group interactions for role classification.
    \item \textbf{AllSet}~\cite{chien2021you} proposes a unified framework that encompasses various hypergraph neural networks as special cases through multiset function learning.
    AllSet formulates hypergraph learning as two multiset functions, namely $f$ for node-to-hyperedge aggregation and $g$ for hyperedge-to-node aggregation, implemented using Deep Sets or Set Transformers.
    \item \textbf{ED-HNN}~\cite{wang2022equivariant} introduces permutation equivariance and disentangled hypergraph operators for hypergraph representation learning.
\end{itemize}
\section{Construction Guidelines for PDMP} \label{app: pdmp}

This section presents the construction guidelines for the \textsc{Pdmp-OD-Det} dataset, a heterogeneous graph derived from Ohio's PDMP data for opioid overdose risk detection.

\subsection{Data Source}
The dataset is sourced from PDMP records provided by the State of Ohio Board of Pharmacy, covering the calendar year 2016 and comprising 1,395,958 controlled substance prescription records.
Each record captures a single dispensing event and includes patient demographics (age, sex, ZIP code), prescriber information (specialty, ZIP code, DEA identifier), pharmacy details (ZIP code, chain affiliation), and medication specifics (drug name, NDC code, therapeutic class, quantity, days supply, payment type).

\subsection{Data Cleaning and Preprocessing}
We apply the following cleaning steps to ensure data quality:
\begin{itemize}[leftmargin=*]
    \item \textbf{Missing Value Handling.} We remove records with missing values in essential fields, including prescription date, patient identifier, prescriber identifier, pharmacy identifier, and therapeutic class code. 
    Moreover, for categorical attributes such as prescriber specialty and ZIP codes, we impute missing values with ``Unknown.''
    \item \textbf{Invalid Record Filtering.} We exclude records with non-positive quantity or days supply values to prevent data entry errors.
    \item \textbf{Date Standardization.} We parse and standardize all prescription dates to ensure temporal consistency across records.
\end{itemize}

\subsection{Entity Construction and Feature Extraction}
We define four node types in the heterogeneous graph. For each entity, we generate textual descriptions from its attributes and encode them using SentenceBERT~\cite{reimers2019sentence} (\texttt{all-mpnet-base-v2}) to produce 768-dimensional feature vectors:
\begin{itemize}[leftmargin=*]
    \item \textbf{Patient (P.)} Unique patient identifiers in the PDMP data constitute patient nodes, which serve as targets for overdose risk prediction. Feature descriptions incorporate patient identifier, age, sex, and ZIP code, e.g., ``ID: [hash], Age: 45, Sex: male, Zip: 43215''.
    \item \textbf{Prescriber (Pr.)} Unique prescribers, such as physicians or healthcare providers, with valid DEA numbers, are represented as prescriber nodes. Feature descriptions incorporate prescriber identifier, ZIP code, specialty, and business activity codes, e.g., ``ID: [hash], Zip: 43215, Specialty: Pain Medicine, BAC: [code]''.
    \item \textbf{Pharmacy (Ph.)} 
    Unique dispensing pharmacies correspond to pharmacy nodes. Feature descriptions include pharmacy identifier, ZIP code, and business activity codes indicating chain affiliation, e.g., ``ID: [hash], Zip: 43210, BAC: [code]''.
    \item \textbf{Drug (D.)} Unique therapeutic class codes define drug nodes. Feature descriptions contain NDC code, drug name, and therapeutic class, e.g., ``NDC: [code], Drug: Oxycodone, Class: Opioid Agonists ([code])''.
\end{itemize}

\subsection{Edge Construction}
The graph contains five edge types derived from prescription records to capture the rich relationships among entities:
\begin{itemize}[leftmargin=*]
    \item \textbf{Patient-take-Drug (P.-take-D.)}: It connects a patient to each therapeutic class of medication they receive. This edge captures the patient's medication profile.
    \item \textbf{Patient-pickup-Pharmacy (P.-pickup-Ph.)}: A patient is linked to each pharmacy where they fill prescriptions. Multiple pharmacy connections may indicate pharmacy shopping behavior.
    \item \textbf{Patient-visit-Prescriber (P.-visit-Pr.)}: It bridges a patient to each prescriber who writes their prescriptions. Multiple prescriber connections may indicate doctor shopping behavior.
    \item \textbf{Prescriber-prescribe-Drug (Pr.-prescribe-D.)}: This denotes a connection between a prescriber and a therapeutic class the prescriber prescribed, capturing prescribing patterns and specialization.
    \item \textbf{Pharmacy-dispense-Drug (Ph.-dispense-D.)} A pharmacy is linked with the therapeutic class it dispenses, indicating the pharmacy-level dispensing patterns.
\end{itemize}
\subsection{Overdose Risk Labeling}
We formulate the opioid overdose detection problem as a binary classification task based on the overdose risk, based on established clinical guidelines~\cite{dowell2022cdc, lossio2022opioid2mme}:
\begin{itemize}[leftmargin=*]
    \item \textbf{Positive Class (High Risk).} Patients whose daily MME exceeds the CDC-recommended threshold, i.e., 90 MME/day, are labeled as high risk. 
    Mention that MME provides a standardized measure of opioid dosage across different medications, enabling consistent risk assessment.
    \item \textbf{Negative Class (Low Risk).} Patients with daily MME below the threshold are labeled as low risk.
\end{itemize}
The resulting heterogeneous graph contains 54,318 nodes and 321,437 edges across five relation types. 
Table~\ref{tab:pdmp_stat} summarizes the data statistics for \textsc{Pdmp-OD-Det}.

\section{Construction Guidelines for X/Twitter} \label{app: x}

In this section, we discuss the data construction and annotation process for the X/Twitter datasets: \textsc{X-HyDrug-Comm}, \textsc{X-HyDrug-Role}, and \textsc{X-MRDrug-Role}.

\subsection{Data Collection}
We first collect metadata via the official X (formerly Twitter) API between December 2020 and August 2021, comprising 275,884,694 posts from 40,780,721 users.
We filter for drug-relevant content using a keyword list of 21 drug types linked to overdose and addiction (proposed by CDC), as listed in Table~\ref{tab:drug_community}, leading to 266,975 drug-relevant posts from 54,680 users for annotation.

\subsection{Annotation Process}
Six domain experts spend 62 days to annotate the filtered users based on their posts, profiles, and interaction patterns.
Each annotator labels users independently, in accordance with the guidelines below.
When annotators disagree on a label, they discuss the case until reaching a consensus.

\subsection{Community Annotation Rules}
For \textsc{X-HyDrug-Comm}, we define six community types by pharmacological function: cannabis, opioid, hallucinogen, stimulant, depressant, and others, and we assign users to one or more communities based on the following rules:
\begin{itemize}[leftmargin=*]
    \item \textbf{Active Promotion.} Users who promote specific drugs on X or interact frequently (following, replying, liking, retweeting) with drug-related accounts in a given community. For example, a seller advertising oxycodone, cocaine, and Xanax would be assigned to the opioid (oxycodone), stimulant (cocaine), and depressant (Xanax) communities.
    \item \textbf{Overdose or Addiction Evidence.} Users who post about overdose or addiction to a specific substance. 
    For example, a user sharing opioid overdose experiences would be placed in the opioid community.
    \item \textbf{Purchase Evidence.} Users with evidence of purchasing specific drugs from others on X. Community assignment follows from the substances purchased.
    \item \textbf{Active Discussion.} Users who discuss or share drug-related content without direct evidence of selling, buying, or using. 
    For instance, A user who frequently retweets LSD promotion will be placed in the hallucinogen community.
    \item \textbf{Unspecified Drugs.} Users involved in drug-related activity who do not mention specific drug types are assigned to the ``others'' community.
\end{itemize}

Since users may engage with multiple drug types, the task is formulated as overlapping community detection that each user can belong to multiple communities.
Table~\ref{tab:drug_community} maps drug types to communities.

\begin{table}[h]
\centering
\caption{Drug community types and drug keywords.}
\label{tab:drug_community}
\resizebox{\linewidth}{!}{
\begin{tabular}{p{2.5cm}p{6cm}}
\toprule
\textbf{Community Type} & \textbf{Representative Drugs (Keywords)} \\
\midrule
Cannabis & Cannabis and cannabis-infused products \\
Opioid & Oxycodone, Codeine, Morphine, Fentanyl, Hydrocodone \\
Hallucinogen & LSD, MDMA, Shroom, Mescaline, DMT, Ketamine \\
Stimulant & Cocaine, Amphetamine, Methamphetamine \\
Depressant & Xanax, Farmapram, Valium, Halcion, Ativan, Klonopin \\
Others & Drugs not listed above \\
\bottomrule
\end{tabular}
}
\end{table}

\subsection{Role Annotation Rules}
For \textsc{X-HyDrug-Role} and \textsc{X-MRDrug-Role}, we classify users into four roles based on their behavior in drug trafficking networks:

\begin{itemize}[leftmargin=*]
    \item \textbf{Drug Seller.} We classify users who advertise or sell drugs on X as drug sellers.
    These accounts typically post drug names, street slang, prices, or contact information, and frequently use drug-related hashtags and emojis to attract buyers.
    \item \textbf{Drug Buyer.} Users with evidence of purchasing drugs on X, including those who respond to advertisements, negotiate transactions, or discuss past purchases, are considered as drug buyers.
    \item \textbf{Drug User.} Users showing signs of drug use, addiction, or overdose belong to the drug users. They may post about consumption habits or withdrawal symptoms, or seek information about drug effects, without evidence of selling or buying.
    \item \textbf{Discussion.} Users who discuss or share drug-related content without evidence of selling, buying, or using are classified as discussion. 
    This includes harm reduction discussions, news sharing, and general commentary.
\end{itemize}

Each user is assigned exactly one role based on dominant behavior, resulting in a multi-class classification task.

\subsection{Feature Extraction}
For each user node, we extract text from the anonymized usernames, profiles, and tweets.
We split tweets into drug-related and drug-unrelated groups using the keyword list.
We keep all drug-related tweets and up to five drug-unrelated tweets per user, favoring those with interactions in the drug community.
We encode the combined text with SentenceBERT~\cite{reimers2019sentence} to produce fixed-length feature vectors.

\subsection{Hyperedge Construction}
For the hypergraph datasets (\textsc{X-HyDrug-Comm} and \textsc{X-HyDrug-Role}), we define the following four hyperedge types:
\begin{itemize}[leftmargin=*]
    \item \textbf{Users-Follow-User} connects users who follow the same account, capturing follower circles within drug communities.
    \item \textbf{Users-Engage-Conversation} links users in the same tweet thread (posting, replying, retweeting, or liking), capturing shared topics.
    \item \textbf{Users-Include-Hashtag} connects users whose tweets contain the same drug-related hashtag.
    \item \textbf{Users-Contain-Emoji} connects users who use the same drug-related emoji in tweets or profiles.
\end{itemize}

\begin{table}[t]
\centering
\caption{Statistics for the benchmark \textsc{PDMP-OD-Det}.}
\label{tab:pdmp_stat}
\setlength{\tabcolsep}{12pt} 
\renewcommand{\arraystretch}{0.75}
\resizebox{\columnwidth}{!}{%
\begin{tabular}{lcc}
\toprule
\textbf{Node Type} & \textbf{\#Nodes} & \textbf{Feature Dim} \\
\midrule
Patient (\textbf{P.}) & 30,574 & 768 \\
Prescriber (\textbf{Pr.}) & 21,159 & 768 \\
Pharmacy (\textbf{Ph.}) & 2,517 & 768 \\
Drug (\textbf{D.}) & 68 & 768 \\
\midrule
\textbf{Edge Type} & \multicolumn{2}{c}{\textbf{\#Edges}} \\
\midrule
\textbf{P.}-take-\textbf{D.} & \multicolumn{2}{c}{76,172} \\
\textbf{P.}-pickup-\textbf{Ph.} & \multicolumn{2}{c}{57,700} \\
\textbf{P.}-visit-\textbf{Pr.} & \multicolumn{2}{c}{86,055} \\
\textbf{Pr.}-prescribe-\textbf{D.} & \multicolumn{2}{c}{66,544} \\
\textbf{Ph.}-dispense-\textbf{D.} & \multicolumn{2}{c}{34,966} \\
\midrule
\textbf{Metapath Type} & \multicolumn{2}{c}{\textbf{\#Edges}} \\
\midrule
\textbf{P.}-\textbf{D.}-\textbf{P.} & \multicolumn{2}{c}{306,338} \\
\textbf{P.}-\textbf{Ph.}-\textbf{P.} & \multicolumn{2}{c}{303,945} \\
\textbf{P.}-\textbf{Pr.}-\textbf{P.} & \multicolumn{2}{c}{265,346} \\
\textbf{P.}-\textbf{Pr.}-\textbf{D.}-\textbf{Pr.}-\textbf{P.} & \multicolumn{2}{c}{305,375} \\
\textbf{P.}-\textbf{Ph.}-\textbf{D.}-\textbf{Ph.}-\textbf{P.} & \multicolumn{2}{c}{305,897} \\
\bottomrule
\end{tabular}%
}
\end{table}

\begin{table}[t]
\centering
\caption{Statistics for the benchmark \textsc{NHANES-Diet}.}
\label{tab:nhance_stat}
\setlength{\tabcolsep}{12pt} 
\renewcommand{\arraystretch}{0.75} 
\resizebox{\columnwidth}{!}{%
\begin{tabular}{lcc}
\toprule
\textbf{Node Type} & \textbf{\#Nodes} & \textbf{Feature Dim} \\
\midrule
User (\textbf{U.}) & 4,826 & 37 \\
Food (\textbf{F.}) & 5,896 & 814 \\
Ingredient (\textbf{I.}) & 2,792 & 768 \\
Category (\textbf{C.}) & 174 & 768 \\
Habit (\textbf{H.}) & 54 & 768 \\
\midrule
\textbf{Edge Type} & \multicolumn{2}{c}{\textbf{\#Edges}} \\
\midrule
\textbf{U.}-eat-\textbf{F.} & \multicolumn{2}{c}{273,934} \\
\textbf{F.}-contain-\textbf{I.} & \multicolumn{2}{c}{38,820} \\
\textbf{F.}-belong-\textbf{C.} & \multicolumn{2}{c}{38,820} \\
\textbf{U.}-has-\textbf{H.} & \multicolumn{2}{c}{93,894} \\
\midrule
\textbf{Metapath Type} & \multicolumn{2}{c}{\textbf{\#Edges}} \\
\midrule
\textbf{U.}-\textbf{F.}-\textbf{U.} & \multicolumn{2}{c}{17,546,430} \\
\textbf{U.}-\textbf{H.}-\textbf{U.} & \multicolumn{2}{c}{23,285,466} \\
\textbf{U.}-\textbf{F.}-\textbf{I.}-\textbf{F.}-\textbf{U.} & \multicolumn{2}{c}{22,485,257} \\
\textbf{U.}-\textbf{F.}-\textbf{C.}-\textbf{F.}-\textbf{U.} & \multicolumn{2}{c}{21,604,607} \\
\bottomrule
\end{tabular}%
}
\end{table}

\section{Construction Guidelines for NHANES} \label{app: nhance}

Next, we discuss the construction guidelines for the heterogeneous graph \textsc{NHANES-Diet}.

\subsection{Data Source}
The dataset is obtained from NHANES~\cite{nhance} and covers the period 2003 to 2020, including dietary intake records, demographic information, and drug-use questionnaires.
Food records in NHANES are encoded using Food and Nutrient Database for Dietary Studies (FNDDS) food codes~\cite{montville2013usda}, a database maintained by the USDA that catalogs food and beverage consumption in the What We Eat In America (WWEIA) survey~\cite{dwyer2003estimation}.

\subsection{Entity Construction and Feature Extraction}
We define five node types in the heterogeneous graph, each with distinct feature representations:
\begin{itemize}[leftmargin=*]
    \item \textbf{User (U.)} Unique NHANES respondents constitute user nodes, which serve as targets for opioid misuse prediction. User features consist of 37 demographic attributes, including age, gender, race, education level, income, and health indicators.
    \item \textbf{Food (F.)} Unique FNDDS food codes correspond to food nodes. Food features comprise 814 dimensions: 46 nutrient values (e.g., calories, protein, fat, vitamins, minerals) concatenated with 768-dimensional BERT embeddings of the food description.
    \item \textbf{Ingredient (I.)} Food ingredients from the FNDDS ingredient database define ingredient nodes that link to their parent food items. Feature representations are 768-dimensional BERT embeddings of ingredient descriptions.
    \item \textbf{Category (C.)} WWEIA food categories represent category nodes that group related foods, such as Milk and Dairy, Meat and Poultry, etc. Feature representations are 768-dimensional BERT embeddings of category descriptions.
    \item \textbf{Habit (H.)} Dietary habit patterns derived from behavioral questionnaires constitute habit nodes. Feature representations are 768-dimensional BERT embeddings of habit descriptions.
\end{itemize}

\subsection{Dietary Habit Annotation}
We compile dietary habit data from various NHANES tables, including Diet Behavior and Consumer Behavior questionnaires.
A team of four domain experts meticulously examines the NHANES data to identify features indicative of dietary habits, such as awareness of a healthy diet or frequency of consuming frozen food.
For each identified feature, we apply a threshold-based annotation strategy:
\begin{itemize}[leftmargin=*]
    \item We select the top 10\% and bottom 10\% of respondents based on their responses.
    \item Users in the top 10\% receive a positive habit tag, while users in the bottom 10\% receive a contrasting habit tag.
\end{itemize}
For example, among respondents to the milk consumption questionnaire, the top 10\% receive the habit ``drink lots of milk,'' while the bottom 10\% receive ``drink little or no milk.''
Through this process, we derive 54 distinct dietary habits that capture meaningful patterns in users' eating behaviors.

\subsection{Edge Construction}
We define four edge types that capture relationships among entities:
\begin{itemize}[leftmargin=*]
    \item \textbf{User-eat-Food (U.-eat-F.)} connects a user to each food item they consume based on their dietary intake records. This edge captures individual food consumption patterns.
    \item \textbf{Food-contain-Ingredient (F.-contain-I.)} connects a food item to its constituent ingredients based on the FNDDS ingredient database. This edge enables reasoning about nutritional composition.
    \item \textbf{Food-belong-Category (F.-belong-C.)} links a food node to its WWEIA food category. This edge captures semantic groupings of related foods.
    \item \textbf{User-has-Habit (U.-has-H.)}: connects a user to their assigned dietary habits based on the annotation process. 
    This edge captures behavioral patterns indicative of dietary preferences.
\end{itemize}

\subsection{Opioid Misuse Labeling}
We formulate opioid misuse prediction as a binary classification task based on criteria commonly employed in public health research~\cite{gu2022prevalence}.
Specifically, a user is labeled as positive (opioid misuse) if either: (i) they have records of using heroin within the past year, or (ii) they have records of using prescription opioid medication continuously for more than 90 days. 
Otherwise, the user is labeled as negative.
The resulting heterogeneous graph contains 13,742 nodes and 445,468 edges across four relation types.
Table~\ref{tab:nhance_stat} summarizes the data statistics for \textsc{NHANES-Diet}.

\section{Additional Experiments} \label{app: efficiency}

\begin{figure*}[t]
 \centering
    \includegraphics[width=\linewidth]{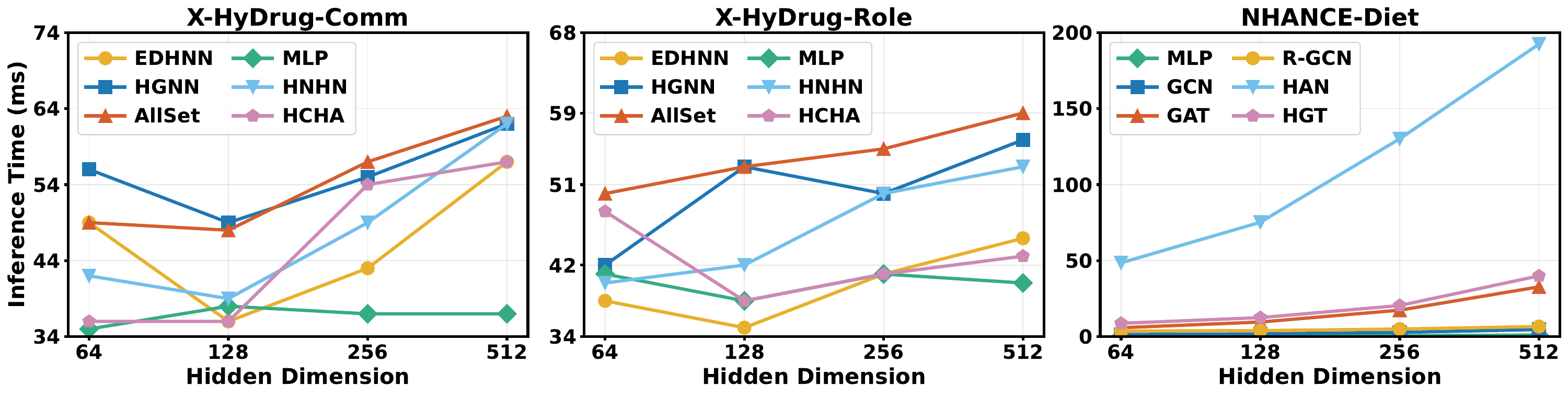}
 \caption{Inference time (ms) across hidden dimensions (64, 128, 256, 512). Left and middle panels show hypergraph neural networks on \textsc{X-HyDrug-Comm} and \textsc{X-HyDrug-Role}; right panel shows heterogeneous graph neural networks on \textsc{NHANES-Diet}.}
 \label{fig: efficiency}
\end{figure*}

To provide practical guidance for deploying graph learning methods on opioid crisis datasets, we conduct an efficiency analysis examining how inference time scales with model capacity.
We evaluate representative methods across \textsc{X-HyDrug-Comm}, \textsc{X-HyDrug-Role}, and \textsc{NHANES-Diet}, varying the hidden dimension from 64 to 512.
Figure~\ref{fig: efficiency} presents the inference time for each configuration.

\noindent\textbf{Illicit Drug Trafficking Detection.}
For hypergraph datasets, MLP exhibits the most stable inference time across all hidden dimensions, ranging from 35 to 38 ms on \textsc{X-HyDrug-Comm} and 38 to 41 ms on \textsc{X-HyDrug-Role}, with minimal sensitivity to increased model capacity.
    Among HyGNNs, HCHA and ED-HNN maintain relatively low inference times even at larger hidden dimensions.
    On \textsc{X-HyDrug-Comm}, ED-HNN ranges from 36 to 57 ms, while HCHA ranges from 36 to 57 ms across dimensions 64 to 512.
HGNN and AllSet show moderate scaling, with inference times increasing from 49 and 56 ms at dimension 64 to 62 and 63 ms at dimension 512.
    HNHN exhibits similar scaling behavior, spanning 42-62 ms.
    On \textsc{X-HyDrug-Role}, all methods demonstrate comparable scaling patterns, with ED-HNN maintaining the lowest inference times and HCHA showing efficient performance.
    
    \noindent\textbf{Opioid Misuse Detection.}
    For the heterogeneous graph dataset \textsc{NHANES-Diet}, we observe substantial variation in inference efficiency across methods.
    MLP is the most efficient, with inference time scaling linearly from 0.22 ms to 1.09 ms.
    GCN provides efficient graph-based inference, with inference times ranging from 1.35 to 4.66 ms across hidden dimensions.
    R-GCN incurs moderate overhead for relation-specific transformations, scaling from 3.65 to 6.64 ms.
    GAT shows quadratic scaling behavior due to attention computation, with inference time increasing from 5.85 ms to 32.75 ms.
    HGT exhibits similar scaling, ranging from 8.76 to 39.96 ms.
    HAN incurs the highest computational cost due to its hierarchical attention mechanism, with inference time scaling from 48.53 ms to 192.45 ms.

\noindent\textbf{Practical Recommendations.}
Based on these findings, we offer the following recommendations for practitioners: 
(1) For hypergraph datasets, all evaluated methods exhibit comparable inference times, making model selection primarily driven by performance requirements rather than computational constraints. 
(2) For heterogeneous graph datasets, attention-based methods (HAN, HGT, GAT) incur significantly higher inference costs that scale poorly with hidden dimension; R-GCN and GCN offer substantially more efficient alternatives suitable for latency-sensitive applications. 
(3) MLP provides the fastest inference across all settings, making it valuable for rapid prototyping or resource-constrained deployments where graph structure can be sacrificed.

\section{Ethical Considerations and Broader Impact}

\subsection{Ethics Statement}
Our research adheres to the ethical standards for data privacy and human subjects protection. 
The PDMP dataset from the State of Ohio Board of Pharmacy is obtained under a data use agreement and is fully de-identified before analysis, ensuring that no personally identifiable information (PII) was exposed. 
Similarly, the NHANES data is a publicly available, de-identified dataset provided by the CDC. 
For the social media data collected from X/Twitter, we strictly follow the platform's Terms of Service and Developer Agreement. 
All user identifiers are hashed, and no individual users are identified in our research including annotation, data construction, and release.
We acknowledge the sensitive nature of opioid-related data and have implemented rigorous security protocols to prevent misuse.
Moreover, we release our benchmark in graph formats that do not contain any raw text or user-level data, further mitigating privacy risks.
Since we use Sentence-BERT~\cite{reimers2019sentence} to generate node features from text data, it is impossible to reconstruct the original text from the features, ensuring that no sensitive information is exposed through the released datasets.

\subsection{Broader Impact}
\modelname has the potential to significantly enhance the capabilities of public health authorities and law enforcement in identifying and intervening in the opioid crisis. 
By providing a standardized platform for graph-based risk detection, we facilitate the development of more accurate early warning systems for overdoses and the disruption of illicit supply chains. 
However, we caution that these models should be used as decision-support tools by qualified professionals to avoid algorithmic bias or stigmatization of vulnerable populations.

\subsection{Limitations}
Despite the comprehensive nature of \modelname, several limitations exist. 
First, the PDMP data is restricted to a single state (Ohio) and a specific timeframe (2016), which may not capture the evolving patterns of the synthetic opioid crisis in recent years. 
Second, online drug trafficking detection on social media platforms only represents a portion of the illicit market, as traffickers frequently migrate to encrypted platforms or the dark web. 
Third, nutritional biomarkers are proxy indicators and should be used as part of a multi-modal assessment rather than a standalone diagnostic tool. 
We will continue maintain and update \modelname to address these limitations by incorporating more recent data, expanding to additional states, and exploring new data sources as they become available.

\begin{table*}[t]
    \centering
    \caption{Best hyperparameters for models on \modelname. LR, WD, Drop, and Dim denote learning rate, weight decay, dropout, and hidden dimension, respectively.
    OOM indicates out-of-memory error during training. 
    } \label{tab: parameters}
    \setlength{\tabcolsep}{3pt}
    \renewcommand{\arraystretch}{0.85} 
    \resizebox{\linewidth}{!}{
    \begin{tabular}{c l cccc cccc cccc}
    \toprule
     & & \multicolumn{4}{c}{Train-10\%Valid-10\%Test-80\%} 
     & \multicolumn{4}{c}{Train-20\%Valid-10\%Test-70\%} 
     & \multicolumn{4}{c}{Train-50\%Valid-10\%Test-40\%} \\
    \cmidrule(lr){3-6} \cmidrule(lr){7-10} \cmidrule(lr){11-14}
     & Model & LR & WD & Drop & Dim & LR & WD & Drop & Dim & LR & WD & Drop & Dim \\
    \midrule\midrule
    \multirow{7}{*}{\rotatebox[origin=c]{90}{\textsc{Pdmp-OD-Det}}} 
    & MLP & 0.001 & 0.0001 & 0.0 & 512 & 0.01 & 0.0 & 0.5 & 512 & 0.01 &  0.0001 & 0.5 & 256\\ 
    & GCN & 0.0001&0.0001&0.3&256&0.0001&0.0&0.5&512&0.0001&0.0&0.5&128\\ 
    & GAT & 0.0001 & 0.0 & 0.5 & 256 & 0.0001 & 0.0001 & 0.3 & 256 & 0.0001 & 0.0 & 0.5 & 256 \\ 
    & R-GCN & 0.01 & 0.0& 0.0 & 512 & 0.0001 & 0.0 & 0.5 & 64 & 0.0001 & 0.0 & 0.5 & 512\\ 
    & HAN & 0.001&0.0001&0.3&512&0.001&0.0&0.3&256&0.001&0.0001&0.3&512\\ 
    & HGT &0.001&0.0&0.0&128&0.001&0.0001&0.3&512&0.0001&0.0001&0.3&128\\ 
    & HGMAE & 0.0001 & 0.0 & 0.5 & 512 & 0.001 & 0.0001 & 0.5 & 512 & 0.01 & 0.0001 & 0.5 & 128 \\ 
    \midrule
    \multirow{8}{*}{\rotatebox[origin=c]{90}{\textsc{X-HyDrug-Comm}}} 
    & MLP & 0.001 & 0.0001 & 0.0 & 256 & 0.0001 & 0.0 & 0.3 & 256 & 0.001 & 0.0001 & 0.3 & 256 \\ 
    & GCN & 0.0001 & 0.0 & 0.3 & 256  & 0.001 & 0.0 & 0.5 & 512 & 0.001 & 0.0001 & 0.3 & 512 \\ 
    & GAT & 0.0001 & 0.0 & 0.3 & 512 & 0.0001 & 0.0005 & 0.3 & 256 & 0.001 & 0.0001 & 0.3 & 256 \\ 
    & HGNN & 0.001 & 0.0005 & 0.3 & 512  & 0.0001 & 0.0 & 0.5 & 512 & 0.0001 & 0.0001 & 0.3 & 512 \\ 
    & HNHN & 0.001 & 0.0001 & 0.3 & 256 & 0.0001 & 0.0 & 0.3 & 256 & 0.001 & 0.0 & 0.3 & 256 \\ 
    & HCHA & 0.0001 & 0.0001 & 0.5 & 256 & 0.0001 & 0.0001 & 0.3 & 256 & 0.001 & 0.0001 & 0.3 & 512 \\
    & AllSet & 0.001 & 0.0 & 0.3 & 128 & 0.0001 &  0.0 & 0.5 & 512 & 0.001 & 0.0 & 0.5 & 256 \\ 
    & ED-HNN & 0.001 & 0.0 & 0.3 & 256 & 0.0001 & 0.0 & 0.5 & 512 & 0.001 & 0.0 & 0.5 & 256 \\ 
    \midrule
    \multirow{8}{*}{\rotatebox[origin=c]{90}{\textsc{X-HyDrug-Role}}} 
    & MLP & 0.0001 & 0.0 & 0.5 & 256 & 0.001 & 0.0 & 0.5 & 256 & 0.001 &0.0 & 0.0 & 256 \\ 
    & GCN & 0.0001& 0.0001 & 0.5& 256& 0.001 & 0.0 & 0.5 & 256 & 0.001 & 0.0 & 0.0 & 512 \\ 
    & GAT & 0.0001& 0.0& 0.3 & 256 & 0.001 & 0.0 & 0.5 & 512 & 0.001 &  0.0 & 0.3 & 256 \\ 
    & HGNN & 0.0001& 0.0001 & 0.5& 512 & 0.001  & 0.0 & 0.5 & 256 & 0.001 & 0.0005 & 0.3 & 512 \\ 
    & HNHN & 0.0001& 0.0& 0.0 & 512 & 0.0001 & 0.0005 & 0.3 &  512& 0.001 &0.0001 & 0.3 & 512 \\ 
    & HCHA & 0.0001& 0.0001 & 0.5 & 256 & 0.0001 & 0.0 &  0.3 & 256  & 0.0001& 0.0001 &  0.5 & 256 \\
    & AllSet & 0.0001& 0.0& 0.3 & 256 & 0.001 & 0.0001 &  0.5 & 256 & 0.001 & 0.0 & 0.5 & 256 \\ 
    & ED-HNN & 0.0001&0.0001  & 0.3 & 512 & 0.001 & 0.0001 & 0.5& 512 & 0.001 & 0.0001 & 0.5 & 512 \\ 
    \midrule
    \multirow{9}{*}{\rotatebox[origin=c]{90}{\textsc{X-MRDrug-Role}}} 
    & MLP & 0.001 & 0.0005& 0.5 & 256 & 0.001 & 0.0005& 0.5 & 256 & 0.001 & 0.0005& 0.5 & 256 \\ 
    & Oversampling & 0.001 & 0.0005& 0.5 & 256 & 0.001 & 0.0005& 0.5 & 256 & 0.001 & 0.0005& 0.5 & 256 \\
    & SMOTE &0.001 & 0.0005& 0.5 & 256 & 0.001 & 0.0005& 0.5 & 256 & 0.001 & 0.0005& 0.5 & 256 \\
    & GCN &0.001 & 0.0& 0.5 & 256 &0.001 & 0.0& 0.5 & 256 &0.001 & 0.0& 0.5 & 256 \\ 
    & GAT &0.001 & 0.0& 0.5 & 256 &0.001 & 0.0& 0.5 & 256 &0.001 & 0.0& 0.5 & 256 \\ 
    & GraphSage &0.001 & 0.0& 0.5 & 256 &0.001 & 0.0& 0.5 & 256 &0.001 & 0.0& 0.5 & 256 \\ 
    & GraphENS &0.001 & 0.0& 0.5 & 256 &0.001 & 0.0& 0.5 & 256 &0.001 & 0.0& 0.5 & 256 \\ 
    & GraphSMOTE &0.001 & 0.0& 0.5 & 256 &0.001 & 0.0& 0.5 & 256 &0.001 & 0.0& 0.5 & 256  \\ 
    & AD-GSMOTE &0.001 & 0.0005& 0.5 & 256&0.001 & 0.0005& 0.5 & 256&0.001 & 0.0005& 0.5 & 256\\ 
    \midrule
    \multirow{6}{*}{\rotatebox[origin=c]{90}{\textsc{NHANES-Diet}}} 
    & MLP & 0.001&0.0 &0.5 &256 & 0.001&0.0 &0.5 &256 & 0.001&0.0 &0.5 &256 \\ 
    & GAT & 0.001&0.0 &0.5 &256 & 0.001&0.0 &0.5 &256 & 0.001&0.0 &0.5 &256 \\
    & GCN & 0.001&0.0 &0.5 &256 &0.001&0.0 &0.5 &256 & 0.001&0.0 &0.5 &256 \\ 
    & R-GCN & 0.001&0.0001 &0.5 &128 & 0.001&0.0001 &0.5 &128 & 0.001&0.0001 &0.5 &128 \\ 
    & HAN & 0.001&0.0 &0.5 &256 & 0.001&0.0 &0.5 &256 & 0.001&0.0 &0.5 &256 \\ 
    & HGT & 0.001 &0.0 &0.5 &128 & 0.001&0.0 &0.5 &128 & 0.001&0.0 &0.5 &128 \\ 
    \bottomrule
    \end{tabular}
    }
\end{table*}

\end{document}